\PassOptionsToPackage{table}{xcolor}
\documentclass[10pt,a4paper,logo]{qwen}
\usepackage{array,float,placeins,wrapfig,multirow}
\usepackage{pdfpages}
\makeatletter
\newcommand{\fs@specalgorithm}{%
  \def\@fs@cfont{\normalfont\small}%
  \let\@fs@capt\floatc@ruled
  \def\@fs@pre{{\color{qwenblue}\hrule height 0.9pt}\kern5pt}%
  \def\@fs@post{\kern5pt{\color{qwenblue!25}\hrule height 0.5pt}}%
  \def\@fs@mid{\kern5pt{\color{qwenblue!25}\hrule height 0.5pt}\kern6pt}%
  \let\@fs@iftopcapt\iftrue}
\makeatother
\floatstyle{specalgorithm}
\newfloat{algorithm}{tbp}{loa}
\floatname{algorithm}{Algorithm}
\floatstyle{plain}
\hypersetup{allcolors=blue,pdftitle={SpecOPD: Branch-Parallel On-Policy Distillation for Speculative Decoding}}
\definecolor{qwenblue}{HTML}{2233FF}

\newtcolorbox{recipecard}[1]{enhanced,breakable,sharp corners,
  colback=qwenblue!2,colframe=qwenblue!25,boxrule=0.5pt,
  borderline north={0.9pt}{0pt}{qwenblue},
  colbacktitle=qwenblue!7,coltitle=black,fonttitle=\bfseries\small,
  title={#1},left=8pt,right=8pt,top=4pt,bottom=4pt,
  before skip=6pt,after skip=7pt}

\newcommand{\method}{\textsc{SpecOPD}}
\newcommand{\E}{\mathbb{E}}
\newcommand{\TV}{\operatorname{TV}}
\newcommand{\code}[1]{\texttt{\small #1}}
\DeclareRobustCommand{\captionstep}[1]{{\def\liningnums{\fontfamily{qhv}\selectfont}\textcircled{#1}}}
\newcommand{\attentionicon}[1]{\raisebox{-0.12em}{\includegraphics[height=0.9em]{figures/attention_#1.pdf}}\hspace{0.35em}}
\DeclareRobustCommand{\captionattentionicon}[1]{\raisebox{-0.12em}{\includegraphics[height=0.9em]{figures/attention_#1.pdf}}}

\title{\method: Branch-Parallel On-Policy Distillation for Speculative Decoding}
\newcommand{\equalcontributionmark}{\textsuperscript{*}}
\newcommand{\correspondingauthormark}{\textsuperscript{\dag}}
\newcommand{\projectleadmark}{\textsuperscript{\ddag}}
\newcommand{\authornotes}{\equalcontributionmark Equal contribution.\quad
  \correspondingauthormark Corresponding author.\quad
  \projectleadmark Project lead.}
\newcommand{\projectlinks}{%
  \href{https://github.com/QwenApp/SpecOPD}{\raisebox{-0.13em}{\includegraphics[height=0.95em]{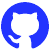}}\hspace{0.35em}Code}%
  \hspace{1.5em}%
  \href{https://qwenapp.github.io/SpecOPD/}{\raisebox{-0.13em}{\includegraphics[height=0.95em]{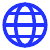}}\hspace{0.35em}Blog}}
\author[1,2]{Tianyu Liu\textsuperscript{*}}
\author[1]{Jun Zhang\textsuperscript{*\ensuremath{\dagger\ddagger}}}
\author[1,3]{Wenjie Huang\textsuperscript{*}}
\author[3]{Yuhao Shen}
\author[1,4]{Tianqi Xu}
\author[1,5]{Xinyue Hao}
\author[1,6]{Hongyaoxing Gu}
\author[7]{Xuyang Liu}
\author[1]{Baolin Zhang}
\author[1]{Yating Zhu}
\author[1]{Shuang Ge}
\author[1]{Jun Dai}
\author[1]{Lei Chen}
\author[1]{Yue Li}
\author[1]{Mingchen Wan}
\affil[1]{Qwen Business Unit of Alibaba}
\affil[2]{University of Science and Technology of China}
\affil[3]{Zhejiang University}
\affil[4]{Carnegie Mellon University}
\affil[5]{University of Edinburgh}
\affil[6]{University of Chinese Academy of Sciences}
\affil[7]{The Hong Kong Polytechnic University}
\begin{abstract}
Speculative decoding (SD) accelerates large language model inference by reducing the sequential cost of token-by-token generation. Advanced SD methods reduce proposal cost with parallel drafters that predict entire blocks in one forward pass, but overall speedup still depends on acceptance length. On-policy distillation (OPD) can push well-trained drafters beyond acceptance-length plateaus, yet conventional pipelines collect supervision through sequential draft-and-verify rollouts.
Our key insight is that on-policy supervision need not inherit these dependencies: a draft block can be supervised using only its reference prefix and draft-sampled history. We propose \method{}, a branch-parallel OPD framework that parallelizes draft sampling and target supervision across independent branches drawn from the evolving drafter policy at reference anchors. We pack these branches into proposal trees to share target-prefix computation, using tree masks for full attention, branch-consistent indexer groups for sparse attention, and indexed state replay for linear attention to preserve each branch's history.
We train the drafter with an E2E-TV surrogate that directly targets acceptance length. Our study spans 8B full-attention (GQA), 27B full--linear (GQA--GDN), and 125B sparse--linear (QSA--GDN) targets, three parallel draft architectures (DFlash, DSpark, and DFlash V2), and four task domains (dialogue, coding, mathematics, and agentic). \method{} achieves up to 41.45\% higher mean acceptance length, $16.09\times$ supervised-token throughput over sequential-rollout OPD under native configurations, and up to $3.56\times$ generation speedup over autoregressive decoding.
\end{abstract}
\renewcommand{\absfont}{\linespread{1.1}\fontsize{10}{12}\selectfont}
\begin{document}
\maketitle

\begingroup
\setlength{\intextsep}{6pt}
\captionsetup{skip=5pt}

\begin{figure}[!ht]
\centering\includegraphics[width=\linewidth]{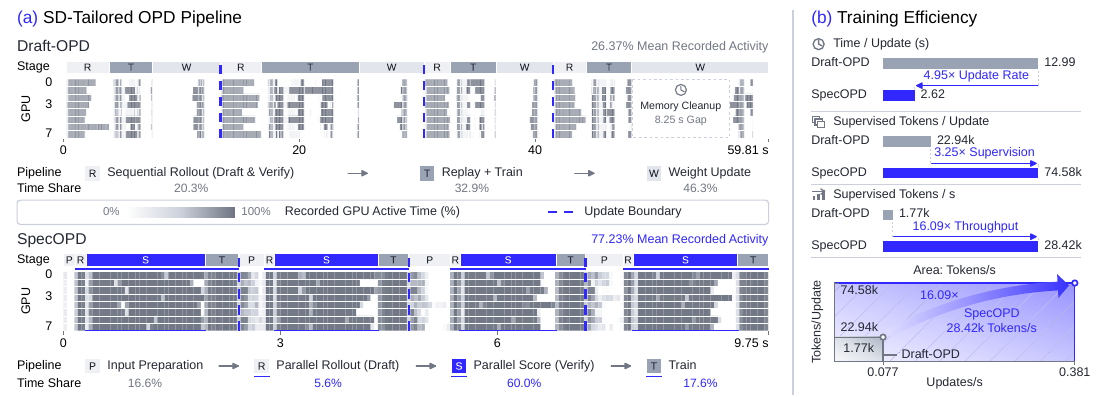}
\caption{\textbf{\method{} improves training throughput.}
Qwen3.8-27B with DFlash V2 on eight high-performance GPUs.
(a) Sequential draft-and-verify rollouts versus parallel drafting and target scoring, shown on separate time axes.
(b) With tokens denoting predictions contributing to the loss, unprofiled measurements show $4.95\times$ update throughput and $3.25\times$ supervised tokens per update, yielding $16.09\times$ supervised-token throughput.}
\label{fig:specopd-efficiency}
\end{figure}

\FloatBarrier
\endgroup
\vspace{-6pt}

\section{Introduction}\label{sec:introduction}
Speculative decoding (SD) accelerates autoregressive generation by verifying multiple tokens proposed by a lightweight drafter in parallel \cite{leviathan,sps2}. Parallel drafters such as DFlash reduce the cost of producing these proposals by predicting an entire block in one forward pass \cite{dflash,peagle,dspark,domino}. The overall speedup also depends on acceptance length, which draft training seeks to improve through better alignment with the target~\cite{distillspec,eagle3}. To this end, a common approach to draft training is supervised fine-tuning (SFT)~\cite{ouyang2022instructgpt} on fixed responses generated by the target model, following sequence-level knowledge distillation~\cite{kim2016seqkd}. On-policy distillation (OPD)~\cite{gkd} has improved acceptance length beyond SFT plateaus by shifting training to blocks sampled from the current draft policy while retaining dense token-level target supervision~\cite{draftopd}.

However, in a typical rollout-and-distill pipeline for SD-tailored OPD, collecting complete trajectories requires sequential draft-and-verify rounds~\cite{draftopd}. The drafter proposes a block, the target verifies it, and the accepted prefix determines where the next round begins. This dependency serializes response collection before drafter updates. 
The measured execution profile in the upper half of Figure~\ref{fig:specopd-efficiency}a shows the alternation between rollout and training, with low GPU utilization during rollout collection.
We therefore ask: \textbf{\textit{Must on-policy training inherit the sequential dependencies of draft-and-verify rollouts?}}

Luckily, these sequential dependencies are not inherent to on-policy supervision. We decouple on-policy supervision from full-response rollouts through a \emph{unique twist} to standard parallel drafter training~\cite{dflash,peagle}: replacing reference-token labels with target supervision on draft-sampled prefixes. We sample anchors from pre-collected responses (regenerated offline by the target or supplied with the raw dataset) and use the current drafter to sample tokens for the masked positions. The target supervises each position using the reference prefix and preceding sampled tokens. Thanks to branch independence across anchors, draft sampling and target supervision can proceed in parallel without sequential full-trajectory collection during training.

To put this parallelism to work, we propose \method, a branch-parallel OPD framework for pushing well-trained drafters beyond acceptance-length plateaus without sequential full-response rollout collection. First, in \textbf{proposal tree construction}, the reference sequence forms a shared \emph{reference trunk}. We randomly sample anchors on this trunk, as in standard parallel-drafter training, and attach continuations sampled from the current drafter as \emph{draft branches}. Second, in \textbf{branch-parallel scoring}, we reuse prefix computation while preserving each branch's history. For full attention  (e.g.,~GQA~\cite{gqa}), we share trunk KV and use tree attention masks to restrict each branch to its reference prefix and own sampled history. For sparse attention (e.g.,~QSA~\cite{qsa}), branch-consistent compressed indexer groups preserve block selection after divergence. For linear attention (e.g.,~GDN~\cite{gdn}), masking alone cannot isolate recurrent states; we therefore use indexed state replay to restore anchor states from shared checkpoints and advance separate branch paths in parallel, avoiding repeated full-prefix execution. Finally, in \textbf{acceptance-length optimization}, we directly target acceptance length with the E2E-TV surrogate~\cite{bebop}, which aggregates prefix products of draft--target distribution overlap along each sampled branch.
The measured execution profile in the lower half of Figure~\ref{fig:specopd-efficiency}a shows the resulting higher GPU utilization.

\noindent To summarize, our main contributions are:

\noindent(1) \textit{Direct, parallel on-policy supervision.} We observe that on-policy supervision for a draft block requires only its reference context and draft-sampled history, without completing a verified response. This yields direct on-policy alignment on draft-generated histories, independently across reference anchors.

\noindent(2) \textit{Novel, efficient SD-tailored OPD.} We propose \method\ to unify independent branch collection, shared-prefix target scoring, and E2E-TV acceptance-length optimization. Its executor preserves branch histories across full (GQA), sparse (QSA), and linear (GDN) attention layers, enabling parallel supervision with prefix reuse.

\noindent(3) \textit{Broad, systematic empirical evaluation.} We evaluate DFlash, DSpark, and DFlash V2 across 8B, 27B, and 125B targets with full, sparse, and linear attention on dialogue, coding, mathematics, and agentic tasks. \method\ improves mean AL by up to 41.45\% and achieves $16.09\times$ supervised-token throughput over sequential-rollout OPD (Figure~\ref{fig:specopd-efficiency}b); measured generation speedup reaches $3.56\times$ over AR.

\section{\method{}}\label{sec:method}
Advanced OPD tailored for SD collects supervision through sequential draft-and-verify rollouts, making each new training context depend on earlier verification outcomes. To remove this collection dependency, \method\ samples independent continuations at reference anchors, scores them through a shared proposal tree, and trains the drafter to increase acceptance length. Let $p$ and $q_\theta$ denote the token distributions of the frozen target and trainable drafter, respectively, and let $x_{1:L}$ be a prompt followed by its pre-collected reference response.

\subsection{Proposal Tree Construction}\label{sec:algorithm}
Sequential rollout collection makes training contexts depend on earlier verification outcomes. We use the reference sequence $x$, including the prompt, as a shared \emph{reference trunk}. Following typical parallel-drafter training~\cite{dflash}, we randomly sample $B$ anchor positions $A=\{a_b\}_{b=1}^{B}$ from its response portion, which can be regenerated offline by the target or supplied with the dataset. At each anchor, we form a block with the clean anchor token in its first position and masks in the remaining positions. The current drafter samples a \emph{draft branch}, a continuation $y_{b,1:W}$ of $W$ tokens excluding the anchor. At position $i=1,\ldots,W$, we define
\begin{equation}
h_{b,i}=(x_{\leq a_b},y_{b,<i}),\qquad
y_{b,i}\sim q_\theta(\cdot\mid h_{b,i}).
\label{eq:branch-history}
\end{equation}
where $h_{b,i}$ combines the reference prefix through the anchor with the preceding samples from the same branch, and $q_\theta$ is the drafter's single-token conditional distribution.

Given the reference sequence $x$ and anchors $A$, branches use independent sampling streams. Their joint distribution $Q_\theta$ over the sampled continuations $Y=\{y_{b,1:W}\}_{b=1}^{B}$ therefore factorizes as
\begin{equation}
Q_\theta(Y\mid x,A)=\prod_{b=1}^{B}\prod_{i=1}^{W}q_\theta(y_{b,i}\mid h_{b,i}).
\label{eq:branch-joint}
\end{equation}
Branches can thus be collected in parallel without waiting for one another's verification outcomes, while allowing dependencies within each branch. On-policy here refers to fresh continuations sampled from the current drafter at fixed reference contexts.

We attach the sampled draft branches $Y$ to their anchors on the reference trunk $x$, forming a \emph{proposal tree}. For branch nodes, the parent map $\pi$ and semantic position are
\begin{equation}
\pi(y_{b,i})=
\begin{cases}
x_{a_b}, & i=1,\\
y_{b,i-1}, & i>1,
\end{cases}
\qquad
\operatorname{pos}(y_{b,i})=a_b+i.
\label{eq:proposal-tree}
\end{equation}
Here, $\operatorname{pos}(\cdot)$ denotes a token's position along its reference prefix and sampled continuation. Trunk nodes satisfy $\pi(x_t)=x_{t-1}$ for $t>1$ and $\operatorname{pos}(x_t)=t$. This representation preserves each branch's reference prefix and own sampled history, exposing shared computation for target scoring.

\begin{figure}[!t]
\centering\includegraphics[width=\linewidth]{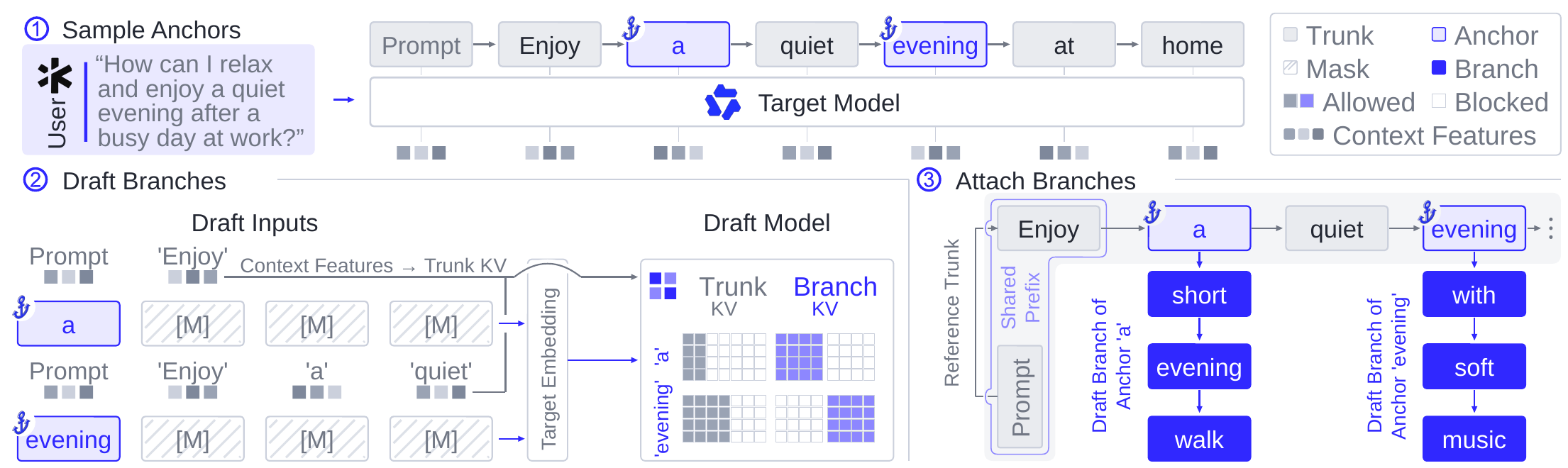}
\caption{\textbf{Proposal tree construction.} \captionstep{1} We randomly sample anchors from a reference response and extract target context features. \captionstep{2} The current drafter samples continuations in parallel at these anchors, conditioned on their respective reference contexts. \captionstep{3} We attach the sampled continuations to their anchors, forming a tree with a shared reference trunk and separate draft branches. Branch independence enables parallel collection, while shared ancestors expose reusable target computation.}\label{fig:construction}
\end{figure}
\FloatBarrier

\subsection{Branch-Parallel Scoring}\label{sec:scoring}
Off-policy training on fixed reference responses aligns the drafter with the target along reference trajectories. To improve acceptance on draft-generated prefixes, \method{} obtains target distributions along continuations sampled from the current drafter and uses them to train the drafter under its own sampled histories. The challenge is to compute these branch-specific distributions in parallel while reusing their shared prefixes. A trunk-only pass misses sampled histories after divergence. Scoring each branch separately recovers the required history, but issuing these short requests sequentially leaves branch parallelism unexploited, while uncached batching repeats overlapping prefix computation.

Our key insight is that distinct supervision histories can still share the computation of their common prefixes. We therefore prefill the reference trunk once and batch branch scoring over its cached representations, reducing redundant computation and amortizing request overhead while preserving the target query $p(\cdot\mid h_{b,i})$. Let $H_u$ be the target hidden state after tree node $u$ and $g_p$ its frozen output head. The readout is
\begin{equation}
u_{b,i}=\pi(y_{b,i}),\qquad
p_{b,i}=p(\cdot\mid h_{b,i})=\operatorname{softmax}\!\left(g_p(H_{u_{b,i}})\right).
\label{eq:tree-readout}
\end{equation}
\begin{wrapfigure}[28]{r}{0.60\textwidth}
\vspace{-\intextsep}
\centering\includegraphics[width=\linewidth]{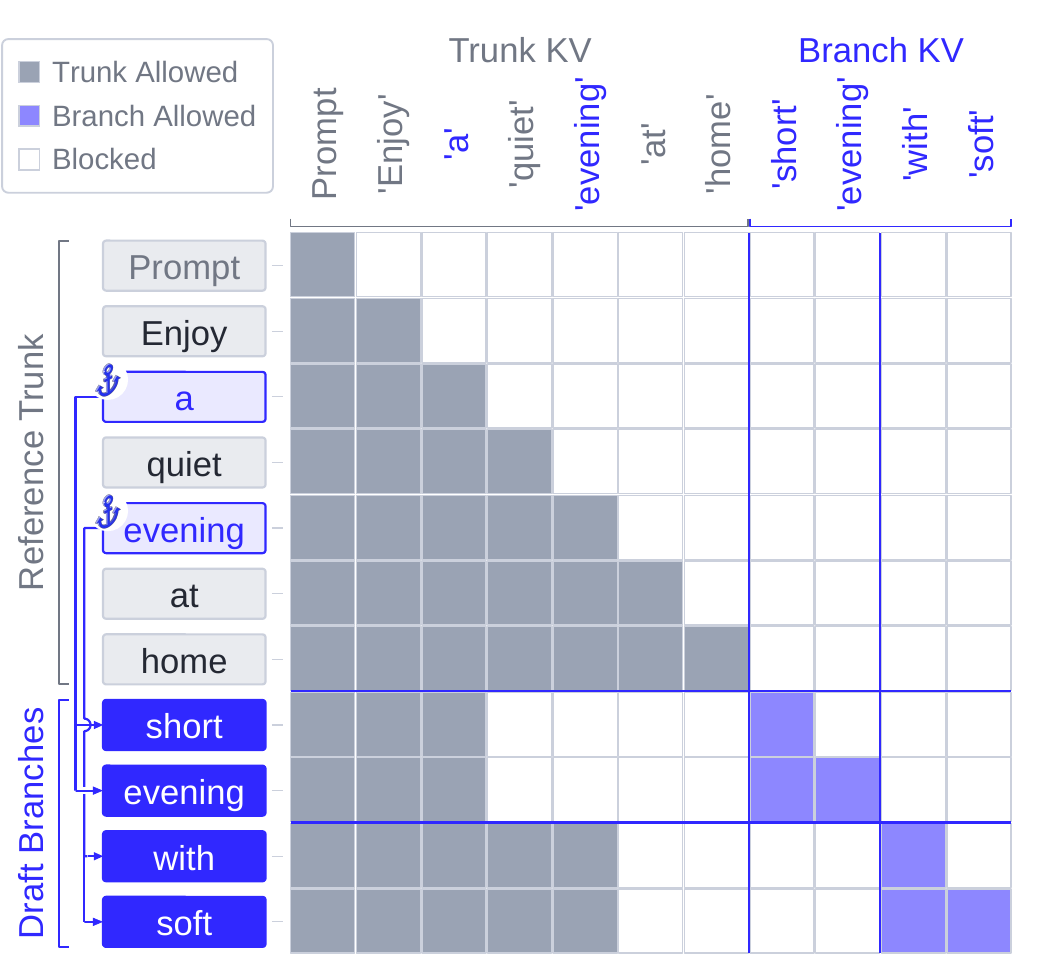}
\caption{\textbf{Branch-parallel scoring with full attention (\captionattentionicon{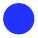}).} The tree attention mask lets packed queries share trunk KV while restricting each row to itself and its ancestors. Rows and columns follow the same packed-token order. Blue arrows connect each anchor to all query rows in its draft branch; \code{Prompt} aggregates the shared prompt prefix. Terminal samples \code{walk} and \code{music} use predecessor readouts and need no query rows.}
\label{fig:tree}
\end{wrapfigure}

The anchor row scores the first token; each subsequent token uses its predecessor row. Thus, $B$ branches contribute $B(W-1)$ new scoring rows after shared trunk processing, with additional state replay where needed. The remaining challenge is to realize this parallelism while preserving each branch's conditioning history across full, sparse, and linear attention.

\paragraph{\texorpdfstring{\attentionicon{full}}{}Full Attention.}
Full attention such as GQA~\cite{gqa} can score sampled branch tokens in parallel once their histories are available. To exploit this parallelism while reusing shared prefixes, we pack branch queries against cached trunk KV. However, a standard causal mask follows the packed order rather than branch ancestry, potentially exposing future trunk tokens or sibling branches. We resolve this mismatch with a tree attention mask, following tree-based speculative verification~\cite{specinfer,medusa}. Let $\operatorname{Anc}(u)$ denote the strict ancestors of $u$ under $\pi$. We use
\begin{equation}
M_{u,v}=
\begin{cases}
0, & v\in\operatorname{Anc}(u)\cup\{u\},\\
-\infty, & \text{otherwise}.
\end{cases}
\label{eq:tree-mask}
\end{equation}
Each query reads only its own path, with rotary positions~\cite{su2021rope} given by $\operatorname{pos}$ rather than packed offsets. Within each attention layer, queries across branches and sampled positions can be evaluated together, reusing trunk KV without re-encoding or replicating shared prefixes. The mask allows each node to attend to itself. This does not leak the token being scored, because Equation~\ref{eq:tree-readout} uses its predecessor's output (Figure~\ref{fig:tree}).

\paragraph{\texorpdfstring{\attentionicon{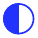}}{}Sparse Attention.}
Sparse attention such as QSA~\cite{qsa} selects token blocks using compressed indexer keys. Recompressing each branch's full context duplicates prefix work, while reusing a cached group crossing an anchor can introduce future reference tokens into block selection. We therefore share complete prefix groups and rebuild the remaining indexer keys in parallel along separate branch histories. Let $\kappa_{b,t}$ denote the raw indexer key at logical position $t$, taken from the reference trunk for $t\leq a_b$ and branch $b$ otherwise. For a query node $u$ on branch $b$, each completed four-token group $g$ with $4g\leq\operatorname{pos}(u)$ uses
\begin{equation}
\bar\kappa_{b,g}=
\begin{cases}
\bar\kappa_g^{\mathrm{trunk}}, & 4g\leq a_b,\\
\mathcal C_g\!\left(\kappa_{b,4g-3:4g}\right), & 4g>a_b.
\end{cases}
\label{eq:qsa-groups}
\end{equation}
Here $g\geq1$, $\bar\kappa_g^{\mathrm{trunk}}$ is a cached trunk group key, and $\mathcal C_g$ applies average pooling, RMSNorm~\cite{zhang2019rmsnorm}, and the target's rotary transform at the group's first position. A four-slot raw-indexer reserve per branch retains the trunk remainder, which combines with branch keys to complete the boundary group before compression. Branch indexer queries score only completed groups on their causal paths, with top-$k$ selection batched and $k$ set by the target's block budget. Attention queries then compute outputs from selected trunk/branch KV and the incomplete causal tail, which bypasses top-$k$. Trunk KV remains shared without prefix replication (Figure~\ref{fig:sparse-groups}). Deterministic selection and traversal are detailed in Appendix~\ref{app:deterministic-target}.

\begin{figure}[!t]
\centering\includegraphics[width=\linewidth]{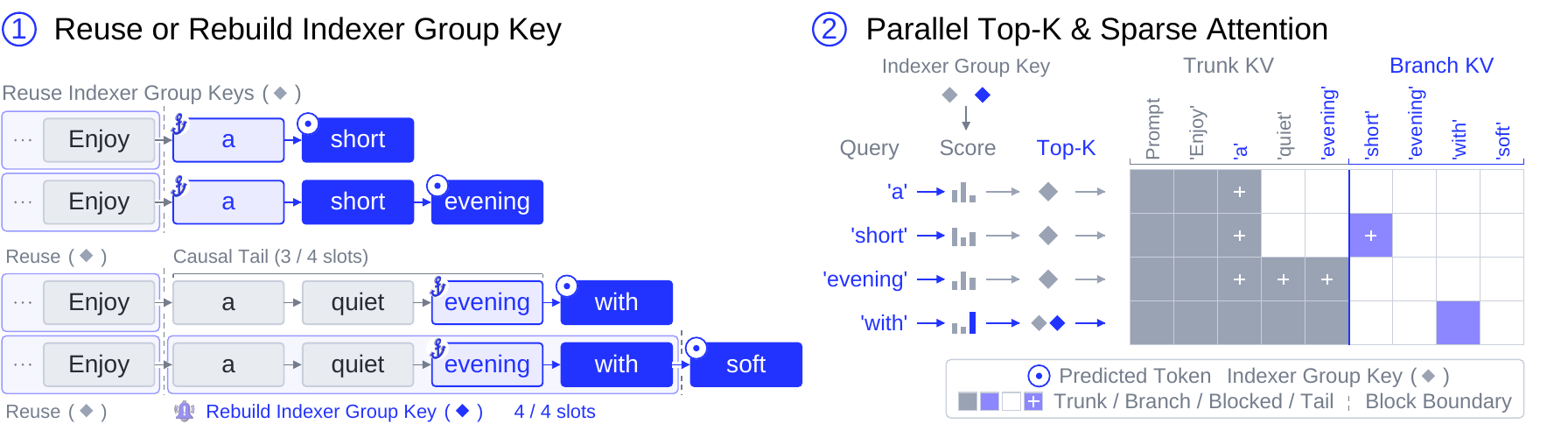}
\caption{\textbf{Branch-parallel scoring with sparse attention (\captionattentionicon{sparse}).} Left-hand tokens are predicted from the predecessor queries named in the right-hand rows. \captionstep{1} All rows reuse complete trunk group keys, including the group ending at \code{Enjoy}. Incomplete boundary groups remain in the causal tail. For the branch at \code{evening}, predicting \code{with} uses the query at \code{evening} and reads \code{[a, quiet, evening]} as a 3/4 causal tail; predicting \code{soft} uses the query at \code{with}, which completes \code{[a, quiet, evening, with]} (4/4) and requires a branch-specific indexer group key. Anchor and bell icons mark branch points and rebuilding. \captionstep{2} Indexer queries select eligible complete groups with top-$k$; attention reads selected KV plus the causal tail, which bypasses top-$k$. }\label{fig:sparse-groups}
\end{figure}

\FloatBarrier
\begin{wrapfigure}{r}{0.6\textwidth}
\vspace{-\intextsep}
\centering\includegraphics[width=\linewidth]{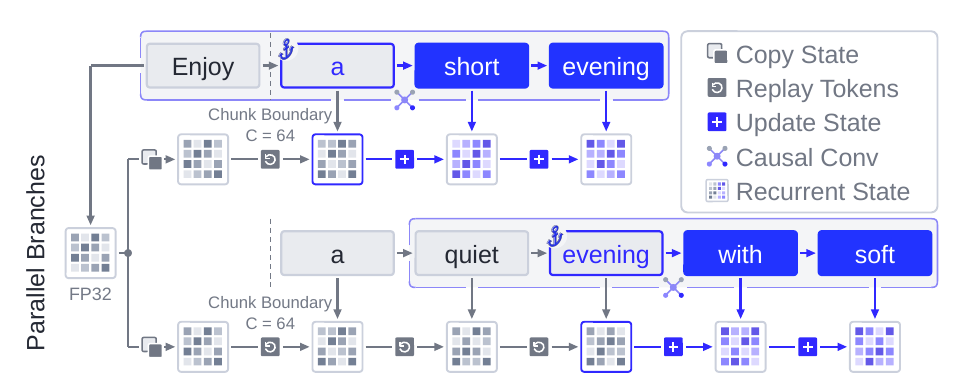}
\caption{\textbf{Branch-parallel scoring with linear attention (\captionattentionicon{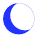}).} A shared trunk pass caches an \texttt{FP32} GDN state after \code{Enjoy} at the $C=64$ boundary (63 prompt tokens omitted). In the indexed recurrent kernel, each thread block (CTA) loads its state tile from this checkpoint into private registers (\emph{Copy State}). \emph{Replay Tokens} uses cached trunk features to advance it through \code{a} or \code{a}$\to$\code{quiet}$\to$\code{evening}, restoring the anchor state. \emph{Update State} then processes branch tokens, updating the state and reading it with each token's query to produce GDN outputs. Checkpoints remain immutable; causal convolution follows branch ancestry.}\label{fig:hybrid-ancestry}
\end{wrapfigure}
\paragraph{\texorpdfstring{\attentionicon{linear}}{}Linear Attention.}
Linear attention such as GDN~\cite{gdn} compresses history into recurrent state, so a mask cannot recover an earlier anchor state or separate subsequent branch updates. The challenge is to restore anchor states cheaply and advance independent recurrences in parallel. We checkpoint the reference trunk and compile each checkpoint-to-anchor suffix and sampled continuation into an indexed path, allowing paths to execute together from separate restored states.

Let $S_t$ denote the GDN recurrent state after reference position $t$, $F$ its one-token update, and $z_t$ the corresponding update features. For branch $b$, replay from the nearest checkpoint $c_b\leq a_b$ restores its anchor state:
\par
\vspace{\baselineskip}
\WFclear
\begin{equation}
\begin{aligned}
S_{b,0}&=\operatorname{Replay}\!\left(S_{c_b};z_{c_b+1:a_b}\right),\\
S_{b,j}&=F(S_{b,j-1},z_{b,j}),\qquad j=1,\ldots,W-1.
\end{aligned}
\label{eq:branch-replay}
\end{equation}

\begin{table}[!t]
\caption{Mean acceptance length ($\uparrow$), including bonus tokens. Darker blue indicates longer Mean AL.}\label{tab:main-acceptance}
\centering\small
\setlength{\tabcolsep}{5pt}
\renewcommand{\arraystretch}{1.10}
\begin{tabular}{cc@{\hspace{14pt}}lcccccc}\toprule
\multicolumn{3}{c}{Setup} & Dialogue & Coding & Math & \multicolumn{2}{c}{Agentic} & \multirow{2}{*}{Average}\\\cmidrule(lr){1-3}\cmidrule(lr){4-4}\cmidrule(lr){5-5}\cmidrule(lr){6-6}\cmidrule(lr){7-8}
Temp. & Target & Draft & MT-Bench & HumanEval & MATH-500 & SWE & Terminal & \\\midrule
& & DFlash  & 2.764 & 3.262 & 3.453 & 2.147 & 1.620 & 2.649\\
\rowcolor{qwenblue} \cellcolor{white} & \cellcolor{white}\multirow{-2}{*}{8B} & \textcolor{white}{\(\hookrightarrow\)\,\method} & \cellcolor{qwenblue!42.6!white}\textcolor{black}{3.432} & \cellcolor{qwenblue!60.3!white}\textcolor{black}{4.177} & \cellcolor{qwenblue!69.7!white}\textcolor{black}{4.571} & \cellcolor{qwenblue!47.0!white}\textcolor{black}{3.617} & \cellcolor{qwenblue!30.8!white}\textcolor{black}{2.940} & \cellcolor{qwenblue!50.1!white}\textcolor{black}{3.747}\\\cmidrule(l){2-9}
& & DSpark  & 3.495 & 3.879 & 4.153 & 3.707 & 3.135 & 3.674\\
\rowcolor{qwenblue} \cellcolor{white} & \cellcolor{white}\multirow{-2}{*}{8B} & \textcolor{white}{\(\hookrightarrow\)\,\method} & \cellcolor{qwenblue!59.0!white}\textcolor{black}{4.121} & \cellcolor{qwenblue!77.0!white}\textcolor{black}{4.876} & \cellcolor{qwenblue!85.3!white}\textcolor{white}{5.223} & \cellcolor{qwenblue!63.3!white}\textcolor{black}{4.301} & \cellcolor{qwenblue!48.4!white}\textcolor{black}{3.676} & \cellcolor{qwenblue!66.6!white}\textcolor{black}{4.439}\\\cmidrule(l){2-9}
& & DFlash V2  & 4.143 & 4.845 & 5.555 & 5.653 & 5.319 & 5.103\\
\rowcolor{qwenblue} \cellcolor{white} & \cellcolor{white}\multirow{-2}{*}{27B} & \textcolor{white}{\(\hookrightarrow\)\,\method} & \cellcolor{qwenblue!63.7!white}\textcolor{black}{4.317} & \cellcolor{qwenblue!81.8!white}\textcolor{white}{5.077} & \cellcolor{qwenblue!100.0!white}\textcolor{white}{5.841} & \cellcolor{qwenblue!95.6!white}\textcolor{white}{5.655} & \cellcolor{qwenblue!87.6!white}\textcolor{white}{5.322} & \cellcolor{qwenblue!85.7!white}\textcolor{white}{5.242}\\\cmidrule(l){2-9}
& & DFlash V2\textsuperscript{\S} & 3.728 & 4.284 & 5.271 & 3.557 & 2.866 & 3.941\\
\rowcolor{qwenblue} \cellcolor{white}\multirow{-8}{*}{\rotatebox{90}{$T=0$}} & \cellcolor{white}\multirow{-2}{*}{125B} & \textcolor{white}{\(\hookrightarrow\)\,\method} & \cellcolor{qwenblue!54.0!white}\textcolor{black}{3.907} & \cellcolor{qwenblue!68.4!white}\textcolor{black}{4.514} & \cellcolor{qwenblue!90.9!white}\textcolor{white}{5.458} & \cellcolor{qwenblue!49.6!white}\textcolor{black}{3.723} & \cellcolor{qwenblue!32.5!white}\textcolor{black}{3.006} & \cellcolor{qwenblue!59.1!white}\textcolor{black}{4.122}\\\midrule
& & DFlash  & 2.521 & 2.930 & 3.269 & 1.891 & 1.596 & 2.441\\
\rowcolor{qwenblue} \cellcolor{white} & \cellcolor{white}\multirow{-2}{*}{8B} & \textcolor{white}{\(\hookrightarrow\)\,\method} & \cellcolor{qwenblue!33.9!white}\textcolor{black}{3.067} & \cellcolor{qwenblue!46.5!white}\textcolor{black}{3.597} & \cellcolor{qwenblue!59.7!white}\textcolor{black}{4.150} & \cellcolor{qwenblue!33.4!white}\textcolor{black}{3.048} & \cellcolor{qwenblue!27.4!white}\textcolor{black}{2.794} & \cellcolor{qwenblue!40.2!white}\textcolor{black}{3.331}\\\cmidrule(l){2-9}
& & DSpark  & 3.330 & 3.641 & 3.997 & 3.197 & 2.929 & 3.419\\
\rowcolor{qwenblue} \cellcolor{white} & \cellcolor{white}\multirow{-2}{*}{8B} & \textcolor{white}{\(\hookrightarrow\)\,\method} & \cellcolor{qwenblue!54.9!white}\textcolor{black}{3.950} & \cellcolor{qwenblue!69.3!white}\textcolor{black}{4.552} & \cellcolor{qwenblue!81.2!white}\textcolor{white}{5.053} & \cellcolor{qwenblue!52.4!white}\textcolor{black}{3.846} & \cellcolor{qwenblue!45.5!white}\textcolor{black}{3.554} & \cellcolor{qwenblue!60.7!white}\textcolor{black}{4.191}\\\cmidrule(l){2-9}
& & DFlash V2  & 3.977 & 4.384 & 5.235 & 5.095 & 4.895 & 4.717\\
\rowcolor{qwenblue} \cellcolor{white} & \cellcolor{white}\multirow{-2}{*}{27B} & \textcolor{white}{\(\hookrightarrow\)\,\method} & \cellcolor{qwenblue!58.9!white}\textcolor{black}{4.115} & \cellcolor{qwenblue!75.7!white}\textcolor{white}{4.820} & \cellcolor{qwenblue!92.1!white}\textcolor{white}{5.511} & \cellcolor{qwenblue!82.7!white}\textcolor{white}{5.116} & \cellcolor{qwenblue!76.3!white}\textcolor{white}{4.845} & \cellcolor{qwenblue!77.1!white}\textcolor{white}{4.881}\\\cmidrule(l){2-9}
& & DFlash V2\textsuperscript{\S} & 3.503 & 3.857 & 4.887 & 3.168 & 2.689 & 3.621\\
\rowcolor{qwenblue} \cellcolor{white}\multirow{-8}{*}{\rotatebox{90}{$T=1$}} & \cellcolor{white}\multirow{-2}{*}{125B} & \textcolor{white}{\(\hookrightarrow\)\,\method} & \cellcolor{qwenblue!47.5!white}\textcolor{black}{3.634} & \cellcolor{qwenblue!56.6!white}\textcolor{black}{4.017} & \cellcolor{qwenblue!82.7!white}\textcolor{white}{5.116} & \cellcolor{qwenblue!41.6!white}\textcolor{black}{3.386} & \cellcolor{qwenblue!29.0!white}\textcolor{black}{2.858} & \cellcolor{qwenblue!51.5!white}\textcolor{black}{3.802}\\\bottomrule
\end{tabular}
\par\vspace{3pt}
{\footnotesize\raggedright \textsuperscript{\S}\,The 125B DFlash V2 baseline was trained in-house on Open-PerfectBlend using LK loss~\cite{lklosses} and D-PACE~\cite{dpace}.\par}
\end{table}
Here $\operatorname{Replay}$ composes $F$ over cached trunk $k,v,g,\beta$ features through the anchor, and $z_{b,j}$ contains the update features of $y_{b,j}$. Each executed branch token's query reads the updated $S_{b,j}$ to produce its GDN output. Replay and continuation updates are batched across branches, while each path retains its causal recurrence. This replaces repeated full-prefix execution with one shared trunk pass and suffix replay from checkpoints. Shared checkpoints remain immutable, and indexed recurrent paths and convolution histories both follow $\pi$ (Figure~\ref{fig:hybrid-ancestry}). Hybrid targets combine this recurrent-state replay with the full or sparse attention mechanisms described above. See Appendices~\ref{sec:gdn}--\ref{app:deterministic-target} for implementation, numerical alignment, and determinism.

\subsection{Acceptance-Length Optimization}\label{sec:acceptance}
\begingroup
\predisplaypenalty=0
We adopt Bebop's E2E-TV objective~\cite{bebop} as an acceptance-length surrogate on draft-sampled branches. At a fixed history, the token acceptance probability under exact rejection sampling equals the draft--target overlap~\cite{leviathan,sps2}:
\begin{equation}
\alpha_{b,i}=\sum_v\min\{q_\theta(v\mid h_{b,i}),p_{b,i}(v)\}
=1-\TV(q_\theta(\cdot\mid h_{b,i}),p_{b,i}),
\label{eq:branch-overlap}
\end{equation}
\endgroup
where $\TV$ denotes total variation distance. For $B$ branches with $W$ proposed tokens each, we minimize
\begin{equation}
\mathcal{L}=1-\frac{1}{BW}\sum_{b=1}^{B}\sum_{j=1}^{W}\prod_{i=1}^{j}\alpha_{b,i}.
\label{eq:loss}
\end{equation}
An early mismatch affects all later prefix terms. We resample branches each update and differentiate draft probabilities and prefix products, holding sampled tokens and target distributions fixed. For conditional proposals, these products are a surrogate rather than an unbiased estimate of serving acceptance (Appendix~\ref{app:toc-subsection-10}).

\FloatBarrier
\section{Experiments}\label{sec:results}

\begin{figure}[!t]
\centering\includegraphics[width=\linewidth]{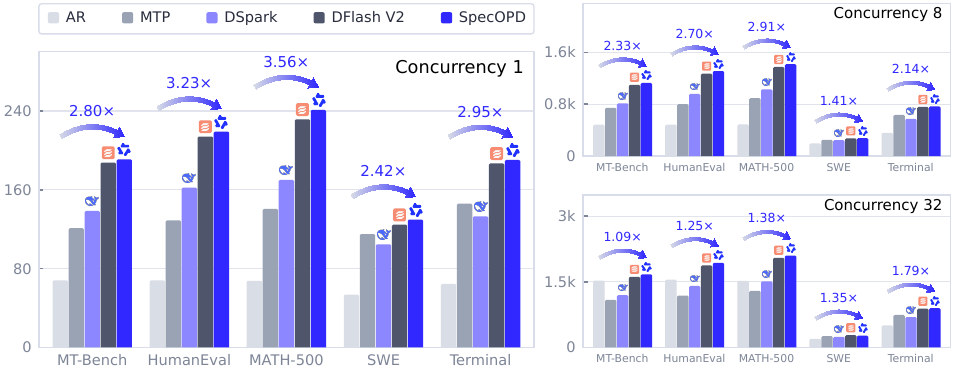}
\caption{Measured Qwen3.8-27B output throughput (tokens/s) with \texttt{BF16}, TP=1, FA3 and CUDA Graphs.}\label{fig:inference-baselines}
\end{figure}

We evaluate \method's inference efficiency, training efficiency, and the role of on-policy learning.

\noindent(1) \textit{Inference efficiency.} Our study spans 8B full-attention, 27B full--linear, and 125B sparse--linear targets; DFlash, DSpark, and DFlash V2; and dialogue, coding, mathematics, and agentic tasks at $T=0$ and $T=1$. Completed comparisons improve five-benchmark mean AL by up to 41.45\% and agentic-task AL by up to 81.51\%. Measured generation speedup reaches $3.56\times$ over AR.

\noindent(2) \textit{Training efficiency.} Against sequential-rollout OPD, \method\ achieves $4.95\times$ higher update throughput while processing $3.25\times$ more supervised tokens per update, yielding $16.09\times$ higher supervised-token throughput. These native-configuration measurements use the 27B target and eight high-performance GPUs and exclude offline reference preparation.

\noindent(3) \textit{Gains from OPD.} At matched optimizer steps and reference-sequence batches, \method\ improves mean AL by up to 15.71\% over off-policy E2E-TV and 6.96\% over off-policy CE. Its best evaluated checkpoint improves prefix survival at all seven draft positions, gaining 4.33\% at position seven, while both off-policy baselines decline from initialization despite decreasing training loss.

\subsection{Setup}\label{sec:setup}
\paragraph{Targets and Training.}
We study Qwen3-8B~\cite{yang2025qwen3} with DFlash and DSpark, Qwen3.8-27B with DFlash V2, and Qwen3.8-Flash-Next-125B with DFlash V2, covering full (GQA), full--linear (GQA--GDN), and sparse--linear (QSA--GDN) attention. We train on Open-PerfectBlend~\cite{openperfectblend} references with fresh draft-sampled branches. Appendix~\ref{app:training-protocol} gives model-specific settings.

\paragraph{Benchmarks and Baselines.}
We evaluate decoding on MT-Bench~\cite{mtbench}, HumanEval~\cite{humaneval}, MATH-500~\cite{math500}, and recorded SWE~\cite{swebench} and Terminal~\cite{terminalbenchpaper} interaction histories. The latter measure continuation efficiency from recorded contexts; Appendix~\ref{app:evaluation-overview} describes the collections. Acceptance comparisons use each drafter's original checkpoint; training comparisons use sequential-rollout OPD, off-policy CE, and off-policy E2E-TV. Serving comparisons include AR, MTP, DSpark and original DFlash V2~\cite{dflash2runtime}.

\paragraph{Evaluation and Metrics.}
Serving evaluations use SGLang~\cite{zheng2024sglang}.
Main comparisons use greedy decoding ($T=0$) and sampling ($T=1$), seven draft tokens, and one selected checkpoint per training line. Acceptance length (AL) counts completion tokens per verification round, including bonus tokens; we average per request and equally across benchmarks. Inference throughput measures output tokens/s. Training efficiency measures optimizer updates/s, supervised tokens/update, and supervised tokens/s; only token predictions contributing to the loss count as supervised tokens. Prefix-survival changes are measured in percentage points. Full metric definitions appear in Appendix~\ref{app:evaluation-overview}.

\subsection{Inference Efficiency}\label{sec:main-results}\label{sec:inference-efficiency}
\paragraph{Acceptance Length across Settings.}
Table~\ref{tab:main-acceptance} shows higher five-benchmark mean AL across all three targets. The largest gains occur on 8B, where DFlash and DSpark improve on every workload at both temperatures; DFlash reaches a 41.45\% mean-AL improvement and an 81.51\% gain on Terminal. Gains are more modest on the hybrid targets, reaching 3.48\% on 27B and 5.01\% on 125B. The 125B drafter improves across all tasks, while 27B agentic results remain nearly unchanged, including a small Terminal decrease at $T=1$. Our 125B drafter was pretrained in-house without agentic data, a coverage gap that may limit adaptation to agentic histories. Extending its pretraining with agentic trajectories is a natural next step to explore further gains.
A separate study starting from public DFlash V2 also shows AL improvements on the \texttt{FP8} Flash-Next target at inference block sizes four and eight (Appendix~\ref{app:public-dflash2-flashnext}).

\paragraph{Throughput across Concurrency.}
Figure~\ref{fig:inference-baselines} reports measured 27B decoding throughput, reaching $3.56\times$ that of AR. Across the five workloads, geometric-mean speedups are $2.97\times$, $2.23\times$, and $1.35\times$ at concurrency 1, 8, and 32. Measured geometric-mean throughput is up to 2.83\% higher than the original DFlash V2 drafter. Gains vary by workload and concurrency; SWE is slower than the original drafter at concurrency 32. Appendix~\ref{app:public-inference} provides the detailed comparison.

\Needspace{5\baselineskip}
\subsection{Training Efficiency}\label{sec:training-efficiency}
\begin{wrapfigure}[44]{r}{0.48\textwidth}
\vspace{-8pt}
\centering
\includegraphics[width=\linewidth]{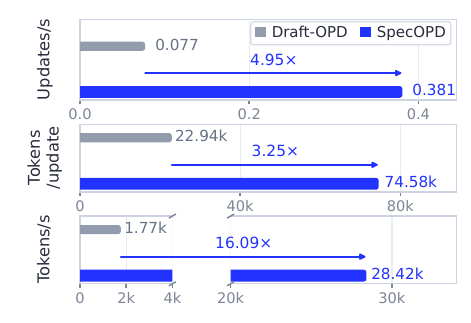}
\caption{Training efficiency on Qwen3.8-27B / DFlash V2. Tokens count supervised predictions contributing to the loss.}\label{fig:training-cost}
\par\vspace{7pt}
\includegraphics[width=\linewidth]{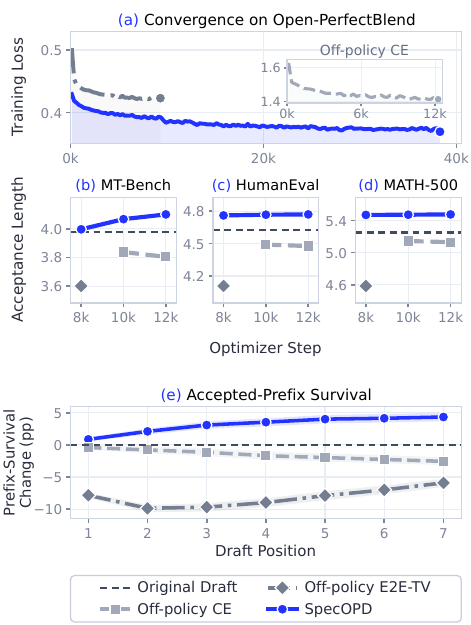}
\caption{Training loss, acceptance length, and prefix survival on Qwen3.8-27B / DFlash V2. Survival changes are relative to initialization; shading shows pointwise 95\% question-cluster bootstrap intervals over 824 paired requests.}\label{fig:offline-losses}
\end{wrapfigure}

\paragraph{Training Throughput.}
On Qwen3.8-27B with DFlash V2 and eight high-performance GPUs, \method{} achieves $16.09\times$ the supervised-token throughput of Draft-OPD (Figure~\ref{fig:training-cost}). This improvement combines $4.95\times$ faster updates with $3.25\times$ more supervised tokens per update: branch parallelism both reduces collection time and supplies more supervised positions at each step.

\paragraph{Execution Profile.}
Figure~\ref{fig:specopd-efficiency}a shows how the collection schedule changes. Draft-OPD alternates dependent rollout, replay/training, and weight-update stages; \method{} batches drafting and target scoring across reference anchors. Mean recorded GPU activity rises from 26.37\% to 77.23\% in the illustrated windows. The profile illustrates how branch parallelism keeps the GPUs active during training. Appendix~\ref{app:throughput-protocol} specifies the timing and profiling procedures.

\subsection{Ablation Study}\label{sec:ablations}

\paragraph{On-Policy vs. Off-Policy.}
We compare \method\ with two off-policy baselines trained on fixed target-generated responses: CE uses reference-token labels, while E2E-TV matches target distributions along reference histories. At matched optimizer steps and reference-sequence batches, the \method{} recipe achieves higher acceptance on all three benchmarks, with a mean advantage of up to 15.71\% over off-policy E2E-TV (Figure~\ref{fig:offline-losses}). \method{} also improves on the common initialization, whereas the evaluated off-policy checkpoints fall below it. The recipes use different compute allocations and gradient paths; Appendix~\ref{app:offline-ablations} gives the configurations and complete comparison.

\paragraph{Training Dynamics.}
Lower training loss does not necessarily translate into longer acceptance. All three runs reduce training loss, but off-policy CE's acceptance declines as training continues, while \method{}'s improves (Figure~\ref{fig:offline-losses}a--d). These trends motivate evaluating acceptance directly alongside the training objective.

\paragraph{Where Acceptance Changes.}
A closer look at prefix survival shows how the gains are distributed across draft positions (Figure~\ref{fig:offline-losses}e), relative to the common initialization. Using each method's best evaluated checkpoint by mean AL, \method{} improves survival across all seven positions, whereas both off-policy baselines show declines. At position seven, survival rises by 4.33 percentage points, compared with a 2.57-point decline for off-policy CE. These request-level measurements connect the training comparison to the accepted-prefix behavior that determines decoding progress.

\Needspace{5\baselineskip}

\section{Related Work}\label{sec:related-work}

\paragraph{Parallel Drafting for SD.}
Speculative decoding (SD) accelerates generation by verifying inexpensive draft proposals with the target in parallel~\cite{leviathan,sps2}. Auxiliary-head and self-speculative approaches improve proposal efficiency through multiple prediction heads or target-layer reuse~\cite{medusa,zhang2024draft,swift,knnssd}. Tree-based verification and confidence-aware candidate allocation improve coverage and budget utilization~\cite{specinfer,eagle2,hu2026echo,liu2026talon}, while retrieval-based and hybrid approaches exploit datastores, cached token transitions, or target logits to provide inexpensive candidates~\cite{rest,tokenrecycling,liu2025logitspec,shen2026graft}. Many drafters reuse target representations: EAGLE drafts autoregressively from target features~\cite{eagle,eagle2,eagle3}, and multi-token prediction (MTP) heads trained with the target serve as built-in drafters~\cite{mtp,deepseekv3}. Autoregressive drafting still spends one drafter pass per proposed token; parallel drafters instead predict a block in one forward pass. Falcon drafts semi-autoregressively~\cite{falcon}, PARD adapts a small draft model to parallel mask prediction~\cite{pard}, and P-EAGLE turns EAGLE into a parallel multi-token predictor~\cite{peagle}. DFlash brings block diffusion to drafting~\cite{dflash}. DFlare enlarges draft capacity with per-layer target-feature fusion~\cite{dflare}. Domino decouples parallel drafting from causal modeling, using a lightweight causal head to refine parallel draft outputs~\cite{domino}; DSpark combines a parallel backbone with lightweight sequential modeling and confidence-scheduled verification~\cite{dspark}. D-cut further adapts verification depth across batched requests to balance acceptance and verification cost~\cite{liu2026dcut}. These designs improve how proposals are generated and verified. \method{} addresses their training: fresh supervision for parallel drafters can be collected across independent reference anchors, without constructing a complete verified trajectory at every collection step. Appendix~\ref{app:detailed-related-work} expands these directions and discusses their connections to RL rollouts and multimodal generation.

\paragraph{On-Policy Draft Training and Acceptance Objectives.}
On-policy distillation supervises student-generated histories~\cite{gkd}. DistillSpec studies draft-generated data and decoding-aware divergence choices~\cite{distillspec}; Draft-OPD uses target-assisted rollouts and error-state replay~\cite{draftopd}. Acceptance objectives account for progress beyond tokenwise accuracy: E2E-TV couples distribution overlap across draft prefixes~\cite{bebop}, while Expected Decoding Rounds models verification-round progress~\cite{edr}. \method{} adopts E2E-TV and parallelizes fresh sampling and history-consistent target scoring, complementing efforts to reduce training memory~\cite{loram} and communication~\cite{edgc}. Its supervision is on-policy within each branch; reference prefixes define the outer contexts and differ from unrestricted full-response rollouts.

\paragraph{Shared Computation across Histories.}
Tree attention, introduced by SpecInfer~\cite{specinfer} and adopted by Medusa~\cite{medusa}, makes multiple candidate paths jointly verifiable while preserving their ancestry. Reusing computation for training must also accommodate targets with different history representations: GQA retains token KV~\cite{gqa}, QSA selects token blocks using compressed indexer keys~\cite{qsa}, and GDN maintains recurrent state~\cite{gdn}. \method{} uses a common proposal tree to organize this reuse, with tree masks, branch-consistent indexer groups, and indexed state replay for the respective attention types. Their common principle is prefix reuse with separate histories after divergence.

\section{Conclusion}\label{sec:conclusion}
\method{} unlocks faster lossless decoding through elegant, efficient branch-parallel on-policy learning built on a simple insight: a draft block can learn directly from target supervision on its own sampled histories. A unique twist on parallel-drafter training replaces reference-token labels with target supervision on draft-sampled histories, while proposal trees enable parallel scoring with shared-prefix reuse. Our study spans 8B full attention (GQA), 27B full--linear (GQA--GDN), and 125B sparse--linear (QSA--GDN) targets, using DFlash, DSpark, and DFlash V2 across dialogue, coding, mathematics, and agentic tasks. \method{} improves mean acceptance length by up to 41.45\% and achieves $16.09\times$ supervised-token throughput over sequential-rollout OPD; measured generation speedup reaches $3.56\times$ over autoregressive decoding. This framework enables the community to explore better policies across heterogeneous targets and translate advances in draft learning into faster production serving.

\section*{Acknowledgments}
We thank Jue Wang and Yubo Wang at \mbox{\raisebox{-0.2em}{\includegraphics[width=1.2em,height=1.2em,keepaspectratio]{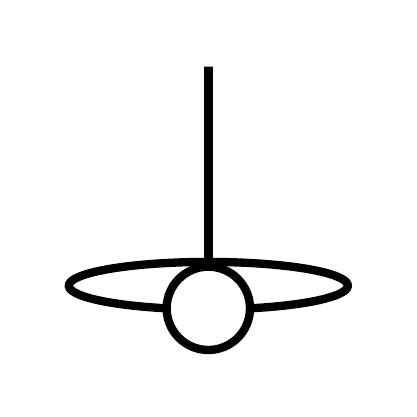}}\,LightSeek Foundation} and the \mbox{\raisebox{-0.03em}{\includegraphics[height=0.72em]{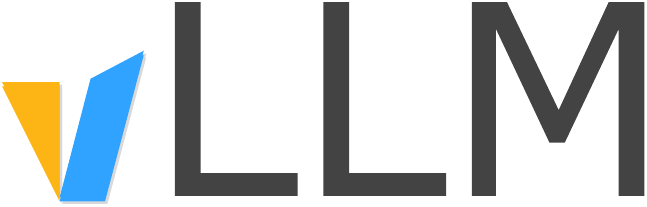}}\ team} for feedback and support.

\clearpage
\appendix
\pdfbookmark[0]{Appendix Contents}{appendix-contents}
\section*{Appendix Contents}
\begingroup
\hypersetup{linkcolor=black}
\small
\newcommand{\appendixtocsection}[2]{\par\addvspace{6pt}\noindent\textbf{\hyperref[#1]{\makebox[3.2em][l]{\ref*{#1}}#2}}\nobreak\dotfill\hyperref[#1]{\pageref*{#1}}\par}
\newcommand{\appendixtocsubsection}[2]{\noindent\hspace*{1.5em}\hyperref[#1]{\makebox[3.2em][l]{\ref*{#1}}#2}\nobreak\dotfill\hyperref[#1]{\pageref*{#1}}\par}

\appendixtocsection{app:supplementary-experiments}{Supplementary Experiments}
\appendixtocsubsection{app:evaluation-overview}{Evaluation Protocols and Metrics}
\appendixtocsubsection{app:8b-protocol}{Qwen3-8B Evaluation}
\appendixtocsubsection{app:27b-evaluation}{Qwen3.8-27B Evaluation}
\appendixtocsubsection{app:125b-hybrid}{Qwen3.8-Flash-Next-125B Evaluation}
\appendixtocsubsection{app:public-dflash2-flashnext}{Public DFlash V2 Training on Flash-Next}
\appendixtocsubsection{app:offline-ablations}{Off-Policy Training Comparisons}
\appendixtocsubsection{app:8b-window}{Inference Window Study on the 8B DFlash Drafter}
\appendixtocsubsection{app:public-inference}{Measured 27B Serving Protocol}
\appendixtocsection{app:training-protocol}{Training Configurations and Measurement}
\appendixtocsubsection{app:training-recipes}{Model-Specific Training Recipes}
\appendixtocsubsection{app:timing}{Training-Throughput Measurement}
\appendixtocsubsection{app:controls}{Comparison Settings}
\appendixtocsection{app:method-details}{Method Details}
\appendixtocsubsection{app:toc-subsection-9}{Conditioning and Prediction-Row Readout}
\appendixtocsubsection{app:toc-subsection-10}{Acceptance Surrogate and Its Scope}
\appendixtocsection{sec:engineering}{Implementation Details}
\appendixtocsubsection{app:toc-subsection-11}{Draft Models}
\appendixtocsubsection{app:toc-subsection-12}{One Optimization Step}
\appendixtocsubsection{sec:gdn}{Parallel Branch Scoring for Hybrid Recurrent Targets}
\appendixtocsubsection{app:deterministic-target}{Deterministic Target Inference}
\appendixtocsubsection{app:toc-subsection-14}{Objective-Preserving Distributed Normalization}
\appendixtocsection{sec:motivation}{Cost and Complexity Analysis}
\appendixtocsubsection{sec:cost}{Collection and Amortized Update Cost}
\appendixtocsubsection{app:toc-subsection-16}{Attention and Recurrent Work}
\appendixtocsubsection{app:toc-subsection-17}{Conditional Wall-Time Sensitivity}
\appendixtocsection{app:detailed-related-work}{Detailed Related Work}

\endgroup
\clearpage
\setlength{\intextsep}{8pt plus 2pt minus 2pt}

\section{Supplementary Experiments}\label{app:supplementary-experiments}
This appendix gives evaluation settings, training trends, and controlled comparisons supporting Section~\ref{sec:results}.

\subsection{Evaluation Protocols and Metrics}\label{app:evaluation-overview}
The five workloads are MT-Bench, HumanEval, MATH-500, SWE, and Terminal. The three standard benchmarks contain 160 MT-Bench turns, 164 HumanEval prompts, and 500 MATH-500 prompts. SWE and Terminal use fixed-history replay: each request continues a recorded assistant-boundary prefix, without live tools or task-success grading. The collections yield 1,662 SWE and 177 Terminal requests, with multiple prefixes possible per conversation. Their membership in the official SWE-bench Verified and Terminal-Bench 2.0 subsets is unconfirmed; the 64 Terminal source trajectories are flagged as overlapping with training data. We use these contexts to measure decoding efficiency.

\paragraph{Checkpoint and Request Matching.}
Each training line uses one checkpoint across all tasks and temperatures in the main comparison. Original and trained weights share valid requests and decoding policies within each study; MT-Bench second turns follow each checkpoint's own first-turn response.

\paragraph{Metrics and Decoding.}
For request $s$, let $t_s$ be completion tokens and $v_s$ verification rounds. AL includes bonus tokens and averages progress per request; the overall result gives equal weight to benchmarks:
\begin{equation}
\overline a_D=\frac{1}{|D|}\sum_{s\in D}\frac{t_s}{v_s},\qquad
\overline a_{\rm avg}=\frac{1}{|\mathcal B|}\sum_{D\in\mathcal B}\overline a_D.
\label{eq:eval-metric}
\end{equation}
Draft-token acceptance pools accepted and proposed tokens, excluding bonus tokens. AL gains are relative changes; prefix-survival changes are measured in percentage points. Output throughput is total final output tokens divided by active client wall time. Main acceptance comparisons propose seven draft tokens with eight verification positions. Greedy decoding uses $T=0$, top-$p=1$, top-$k=1$; sampling uses $T=1$, top-$p=0.95$, top-$k=20$. Serving measurements use one complete run per condition.

For training efficiency, let $U$ be optimizer updates, $N$ supervised tokens, and $t$ elapsed training time. We report updates/s ($U/t$), supervised tokens/update ($N/U$), and supervised tokens/s ($N/t$), so supervised-token throughput equals update throughput multiplied by supervision per update. Only predictions contributing to the loss count toward $N$; context-only and masked-out positions are excluded. Timings include each method's training workflow and exclude offline reference preparation (Appendix~\ref{app:throughput-protocol}).

\subsection{Qwen3-8B Evaluation}\label{app:8b-protocol}
\begin{figure}[!htbp]
\centering\includegraphics[width=\linewidth]{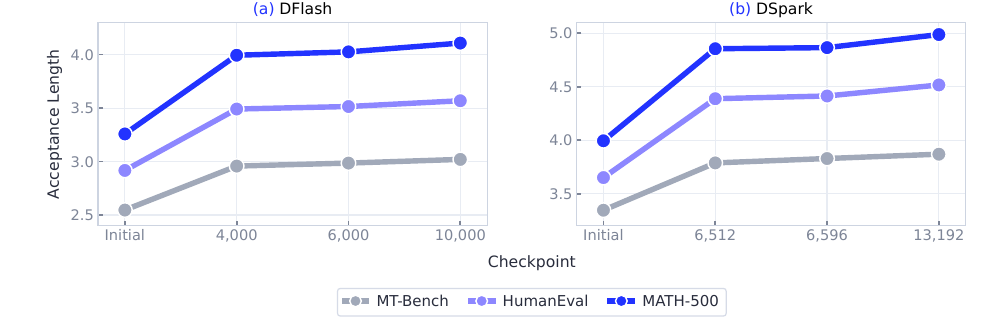}
\caption{Qwen3-8B acceptance during training at $T=1$ and concurrency four. Both drafters propose seven tokens with eight verification positions. Checkpoints are equally spaced; lines connect observed checkpoints. Panels use different y-axis ranges.}\label{fig:8b-checkpoints}
\end{figure}

The main comparison uses DFlash step 20,000 and DSpark step 26,382. Both improve AL and draft-token acceptance over initialization on every task at both temperatures. Figure~\ref{fig:8b-checkpoints} shows their earlier training trends, ending at DFlash 10,000 and DSpark 13,192; Table~\ref{tab:main-acceptance} reports the later selected checkpoints.

The DFlash initialization is \code{z-lab/Qwen3-8B-DFlash-b16}; pretraining and \method\  continuation use block size 16. Evaluation uses block size eight for DFlash and seven for DSpark, matching their seven-token proposal budget. Appendix~\ref{app:8b-window} examines DFlash at both inference windows.

All 8B runs use \texttt{BF16}, TP=1, and one high-performance GPU per service. Standard-task evaluation uses FlashInfer, thinking mode, concurrency four, an 8,192-token context, and a 4,096-token output cap, with CUDA Graphs, radix caching, and overlap scheduling disabled. SWE and Terminal use concurrency 16, a 32,768-token context, and an 8,192-token output cap. Paired filtering retains 1,661 SWE and all 177 Terminal requests. Generation seeds are fixed across compared weights.\label{app:8b-workload-audit}

\subsection{Qwen3.8-27B Evaluation}\label{app:27b-evaluation}
\paragraph{Agentic Workloads.}\label{app:27b-matched-replay}
The main comparison uses step 36,000, selected by standard-task AL, on all 1,662 SWE and 177 Terminal requests. Paired runs use the same GPU pair, TP=2, concurrency 16, a 40,960-token context, an 8,192-token output cap, and token capacity 131,072. The runtime uses an \texttt{FP32} output head, a shared one-stage full-attention decode kernel, and GDN updates without floating-point fusion. Request order and generation seeds match across weights. Complete greedy outputs match across the compared weights on these workloads. At $T=1$, AL increases by 0.40\% on SWE and decreases by 1.02\% on Terminal.

\paragraph{Standard Tasks.}\label{app:27b-t0}
The main-table $T=0$ results use the same step-36,000 checkpoint and the standard request counts listed above. Three-task mean AL rises from 4.8477 to 5.0784 (+4.76\%). These archived runs use eight TP=1 replicas with concurrency eight per replica. Greedy outputs differ between original and trained weights in this archive; the reported comparison measures acceptance. The serving-throughput experiment uses the separate configuration in Appendix~\ref{app:public-inference}.

\subsection{Qwen3.8-Flash-Next-125B Evaluation}\label{app:125b-hybrid}
\begin{figure}[!htbp]
\centering\includegraphics[width=\linewidth]{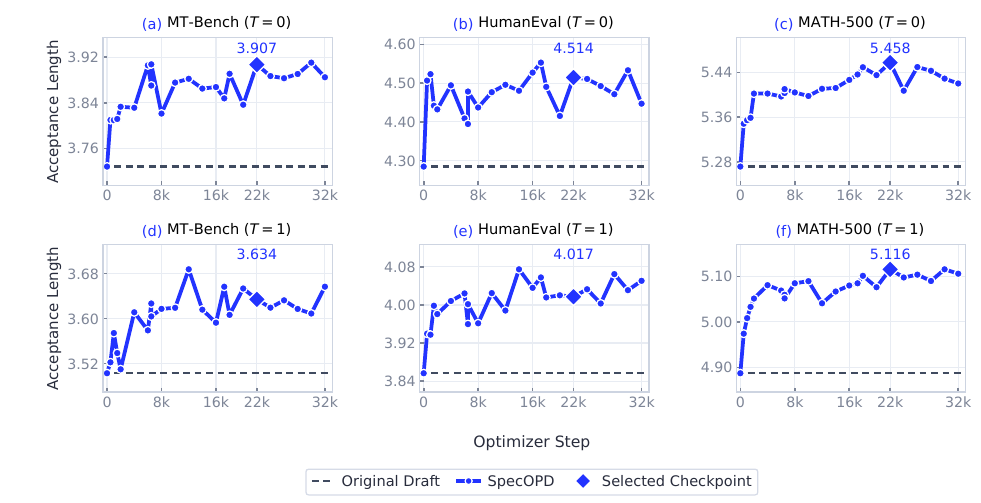}
\caption{125B acceptance during training at concurrency 16. Rows show $T=0$ and $T=1$. The selected checkpoint is step 22,000; lines connect observations through step 32,000.}\label{fig:125b-c16-checkpoints}
\end{figure}

The target contains 36 GDN layers and 12 QSA layers. QSA uses selection budget 2,048, compression ratio four, four indexer heads, and head dimension 128. DFlash V2 initialization is step 196,000; \method\  uses global batch 64, microbatch one, and accumulation four. Figure~\ref{fig:125b-c16-checkpoints} shows the standard-task training trajectory. Step 22,000 maximizes the six-cell mean AL (4.4410) and supplies every 125B trained entry in Table~\ref{tab:main-acceptance}.

Evaluation uses an \texttt{FP8} target and \texttt{BF16} drafter, TP=2/EP=2, concurrency 16, and \code{xhigh} reasoning effort on high-performance GPUs. Standard tasks use a 32,768-token context and 4,096-token output cap; SWE and Terminal use a 40,960-token context and 8,192-token output cap. The runtime uses Triton attention, FA2 for QSA variable-length execution, \texttt{FP32} recurrent state, and disabled CUDA Graphs. Compared weights share the same requests and decoding policies.

\subsection{Public DFlash V2 Training on Flash-Next}\label{app:public-dflash2-flashnext}
\begin{figure}[!htbp]
\centering
\includegraphics[width=\linewidth]{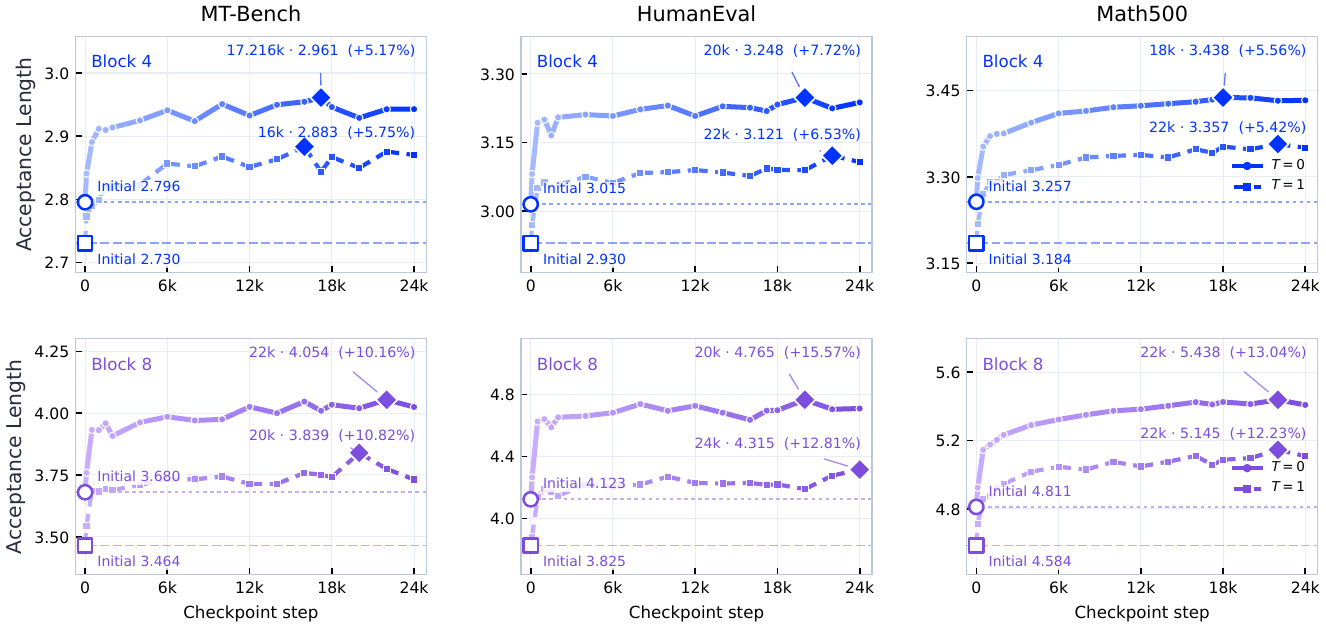}
\caption{Public DFlash V2 acceptance during \method\ training with an \texttt{FP8} Flash-Next target. Rows use inference block sizes four and eight. Diamonds mark each curve's highest observed AL; percentages use that checkpoint's matched initialization. Initial points and horizontal references use the baselines paired with step 22,000. Lines guide the eye across GPU/configuration changes rather than depict a fixed-configuration trajectory.}\label{fig:public-dflash2-flashnext}
\end{figure}

We additionally evaluate \method\ continuation from the public \texttt{BF16} DFlash V2 initialization with an \texttt{FP8} Flash-Next target at inference block sizes four and eight. This study is separate from the in-house initialization used for the 125B main-table comparison. Figure~\ref{fig:public-dflash2-flashnext} covers 17 trained checkpoints per block, from step 100 to 24,000, on MT-Bench, HumanEval, and MATH-500 at both temperatures. The largest observed task--temperature AL improvements over matched initializations are 7.72\% at block four and 15.57\% at block eight; the best checkpoint varies across tasks and temperatures.

Evaluation uses high-performance GPUs with server and request seeds both set to 42. Each checkpoint covers 160 MT-Bench turns (including second turns conditioned on its own first-turn responses), 164 HumanEval requests, and 500 MATH-500 requests at each temperature, for 1,648 responses before filtering. Evaluation outputs flagged as invalid are excluded from aggregation. Each trained checkpoint is compared with the initial model under the same hardware and inference settings; the reported gains use these matched comparisons.

\subsection{Off-Policy Training Comparisons}\label{app:offline-ablations}
Table~\ref{tab:offline-ablations} compares \method\  with off-policy CE and E2E-TV at matched optimizer steps and reference-sequence batches. \method\  improves on the common initialization, while the evaluated off-policy checkpoints fall below it.

\begin{table}[!htbp]
\caption{On-policy and off-policy training on Qwen3.8-27B / DFlash V2 at matched optimizer steps and reference-sequence batches. AL includes bonus tokens ($\uparrow$).}\label{tab:offline-ablations}
\centering\small
\begin{tabular}{llrrrr}\toprule
Steps & Training method & MT-Bench & HumanEval & MATH-500 & Average\\\midrule
0 & Original draft & 3.9787 & 4.6225 & 5.2519 & 4.6177\\\midrule
8k & Off-policy E2E-TV & 3.6011 & 4.1086 & 4.5854 & 4.0983\\
\rowcolor[HTML]{F1F2FF} 8k & \method\  & 3.9966 & 4.7602 & 5.4699 & 4.7422\\\midrule
10k & Off-policy CE & 3.8377 & 4.4905 & 5.1448 & 4.4910\\
\rowcolor[HTML]{F1F2FF} 10k & \method\  & 4.0675 & 4.7668 & 5.4726 & 4.7690\\\midrule
12k & Off-policy CE & 3.8042 & 4.4771 & 5.1311 & 4.4708\\
\rowcolor[HTML]{F1F2FF} 12k & \method\  & 4.1005 & 4.7696 & 5.4764 & 4.7822\\\bottomrule
\end{tabular}
\end{table}

The three training recipes start from public DFlash V2, use fresh optimizers, global sequence batch 96, and peak learning rate $5\times10^{-5}$. Off-policy runs train on fixed target-generated responses with 16 high-performance GPUs (six draft ranks and ten target workers); \method\  uses fresh sampled branches with 32 high-performance GPUs (12 draft ranks and 20 target workers) and merged accumulation. The off-policy objectives follow unary CE/E2E-TV paths, while \method\  differentiates selector-conditioned probabilities. These recipe-level comparisons match sequence presentations, with different compute allocations and gradient paths.

Evaluation uses $T=1$, top-$p=0.95$, top-$k=20$, a 4,096-token completion cap, and concurrency eight per replica. Off-policy E2E-TV is evaluated at 8k steps and CE at 10k and 12k. The prefix-survival figure pairs all 824 requests with the common initialization and excludes bonus tokens. Its 2,000 bootstrap replicates are stratified by task and clustered by question, keeping MT-Bench turns together. The resulting pointwise 95\% intervals describe request-level variation. Benchmark means in the table have equal weights; the position figure gives each request equal weight.

\subsection{Inference Window Study on the 8B DFlash Drafter}\label{app:8b-window}\label{app:8b-window-ablation}
DFlash pretraining and \method\  continuation use block size 16, while the main table uses inference block size eight to match DSpark's seven-token proposal budget. We compare the block-16 initialization with its step-10,000 continuation at inference windows eight and 16 on the three standard tasks at both temperatures. Window eight proposes seven tokens; window 16 proposes fifteen. The four arms use the same high-performance GPU, TP=1, concurrency four, an 8,192-token context, and a 4,096-token output cap, with CUDA Graphs, radix caching, and overlap scheduling disabled.

\begin{figure}[!htbp]
\centering\includegraphics[width=\linewidth]{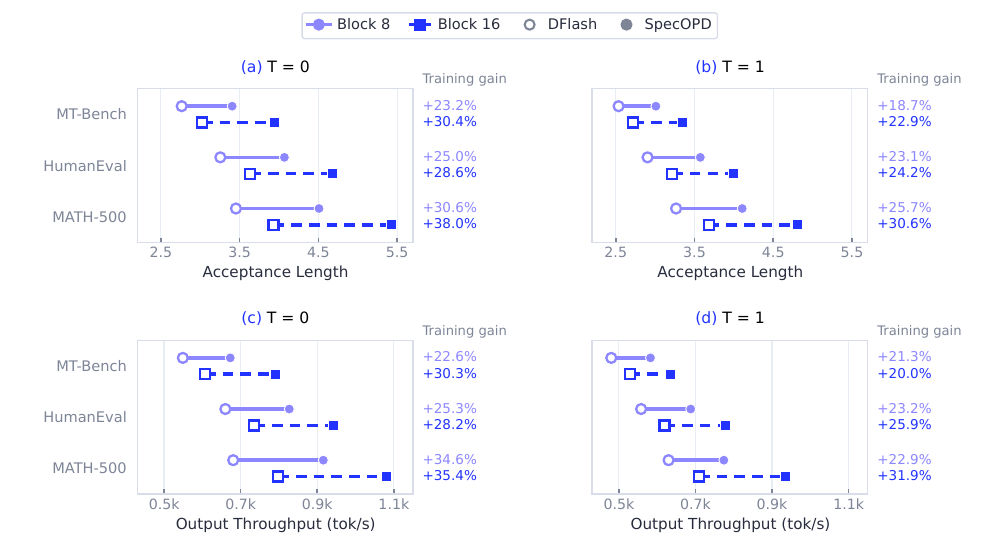}
\caption{Training gains at both DFlash inference windows. \method\  uses the step-10,000 checkpoint; B8/B16 denote inference block size. Columns show $T=0$/$T=1$; rows show AL and throughput. Percentages compare training at a fixed window. All weights are trained with block size 16.}\label{fig:8b-window-training}
\end{figure}

Training improves AL and throughput in all task--temperature pairs at both windows (Figure~\ref{fig:8b-window-training}). AL increases by 18.67--30.57\% at window eight and 22.87--37.97\% at window 16; the corresponding throughput gains are 21.33--34.58\% and 20.04--35.41\%.

\paragraph{Common-Anchor Comparison.}
To compare draft quality at the same proposal budget, we also measure first-verification acceptance at matched prefixes and anchors. Prefixes come from initialization answers at offsets 0, 256, and 1,024 (clipped and deduplicated), yielding 480 MT-Bench, 492 HumanEval, and 1,500 MATH-500 prefixes. Every arm uses the same prefix, seed, and reference anchor from the initialization/window-eight arm; these probes are separate from the throughput runs. For accepted draft tokens $A$, excluding the bonus token, the comparable first-seven score is $S_7=\sum_{j=1}^{7}\Pr(A\geq j)=\mathbb E[\min(A,7)]$.

\begin{table}[!htbp]
\caption{Common-anchor effects on $S_7$, in accepted draft tokens. $\Delta_W$ is window 16 minus 8; $\Delta_{\rm train}$ is step 10,000 minus initialization. Interaction is $\Delta_W(10{,}000)-\Delta_W(0)$.}\label{tab:8b-window-effects}
\centering\small\setlength{\tabcolsep}{4pt}
\begin{tabular}{llrrrrr}\toprule
$T$ & Task & $\Delta_W(0)$ & $\Delta_W(10{,}000)$ & $\Delta_{\rm train}(8)$ & $\Delta_{\rm train}(16)$ & Interaction\\\midrule
0 & MT-Bench & 0.0604 & 0.6896 & 1.6417 & 2.2708 & 0.6292\\
0 & HumanEval & 0.3455 & 0.6524 & 1.9004 & 2.2073 & 0.3069\\
0 & MATH-500 & 0.0780 & 0.4560 & 2.3867 & 2.7647 & 0.3780\\
1 & MT-Bench & 0.0729 & 0.5708 & 1.5312 & 2.0292 & 0.4979\\
1 & HumanEval & 0.2866 & 0.6748 & 1.6992 & 2.0874 & 0.3882\\
1 & MATH-500 & 0.0973 & 0.3167 & 2.2967 & 2.5160 & 0.2193\\
\bottomrule\end{tabular}
\end{table}

Training raises $S_7$ at both windows. Window 16's advantage grows from 0.0604--0.3455 at initialization to 0.3167--0.6896 after training, giving positive interactions in all six conditions (Table~\ref{tab:8b-window-effects}). Both sets of weights are trained with window 16; this study varies the inference window and checkpoint.

\subsection{Measured 27B Serving Protocol}\label{app:public-inference}
Figure~\ref{fig:inference-baselines} compares AR, MTP, DSpark, original DFlash V2, and \method\  at concurrency 1, 8, and 32. The \method\  checkpoint is step 36,000, selected by standard-task AL before this study. Methods within each workload/concurrency group use the same high-performance GPU and CPU affinity; separate groups use separate GPUs and disjoint CPU affinities.

Each method receives the same shuffled input list and seed 42 at a given concurrency: 128 requests at concurrency one and 1,024 at concurrency eight or 32. Short datasets repeat in a fixed order, and MT-Bench uses first turns. SGLang uses \texttt{BF16} target/draft weights and LM head, TP=1, FA3 attention, CUDA Graphs, $T=1$, top-$p=0.95$, top-$k=20$, and \code{xhigh} reasoning effort. The completion cap is 4,096, context length 40,960, KV capacity 131,072, chunked prefill size 2,048, and maximum running requests 32. Speculative methods propose seven tokens with eight verification positions.

The radix cache is flushed after warmup. Timing includes request submission, HTTP, generation, and response receipt; it excludes loading, warmup, tokenizer initialization, and artifact writing. Terminal at concurrency 32 records three KV-cache retractions for AR and two for MTP, whose scheduling costs are included. Each condition is one complete run. This TP=1/\texttt{BF16}-head protocol differs from the TP=2/\texttt{FP32}-head main acceptance study.

Across the five workloads, \method\ 's geometric-mean speedups over AR are $2.97\times$, $2.23\times$, and $1.35\times$ at concurrency 1, 8, and 32, versus $2.89\times$, $2.19\times$, and $1.34\times$ for original DFlash V2. Measured geometric-mean throughput is 2.83\%, 2.20\%, and 1.21\% higher, respectively. SWE at concurrency 32 is 4.46\% slower than the original drafter.

\FloatBarrier

\section{Training Configurations and Measurement}\label{app:training-protocol}
\paragraph{Data and Training.}
We train on Open-PerfectBlend~\cite{openperfectblend}, using reference responses as outer contexts and resampling draft branches at each update. The reference-response source, retained subset, and exact checkpoint identifiers remain to be documented for each run. The cards below summarize the model-specific training settings; the separate offline and training-throughput studies retain their own configurations.

\Needspace{6\baselineskip}
\subsection{Model-Specific Training Recipes}\label{app:training-recipes}
Evaluation budgets and checkpoint selection are specified separately in Appendices~\ref{app:8b-protocol}, \ref{app:27b-evaluation}, and~\ref{app:125b-hybrid}.

\begin{recipecard}{Qwen3-8B | Training Recipe}
\small
\begin{tabularx}{\linewidth}{@{}>{\bfseries}p{0.20\linewidth}X@{}}
Initialization & Public DFlash and DSpark drafters; DFlash: \code{z-lab/Qwen3-8B-DFlash-b16}.\\[1.5pt]
Training window & DFlash block size 16, matching pretraining.
\end{tabularx}
\end{recipecard}

\begin{recipecard}{Qwen3.8-27B | Training Recipe}
\phantomsection\label{app:archived}
\small
\begin{tabularx}{\linewidth}{@{}>{\bfseries}p{0.20\linewidth}X@{}}
Initialization & Public DFlash V2; fresh optimizer; train backbone and conditional selector.\\[1.5pt]
Objective & Pure \method{}, weight 1; at most 512 scored anchors.\\[1.5pt]
Sampling & Proposal / score temperature 1 / 1; selector top-$k$ 16.\\[1.5pt]
Optimizer & AdamW; peak / minimum learning rate $10^{-5}$ / $10^{-6}$; betas $(0.9,0.95)$; weight decay 0.1; gradient-norm limit 1.\\[1.5pt]
Batch & Global batch 48 reference sequences; microbatch 1; accumulation 8.\\[1.5pt]
Execution & Six draft ranks; merged accumulation and step balancing; one concurrent training batch.\\[1.5pt]
Reference data & 1,227,197 filtered sequences, regenerated before training at $T=1$, top-$p=0.95$, top-$k=20$ (distinct from selector support).\\[1.5pt]
Warmup & 7,669 updates.
\end{tabularx}
\end{recipecard}

\begin{recipecard}{Qwen3.8-Flash-Next-125B | Training Recipe}
\small
\begin{tabularx}{\linewidth}{@{}>{\bfseries}p{0.20\linewidth}X@{}}
Initialization & DFlash V2, pretrained in-house; step 196,000.\\[1.5pt]
Batch & Global batch 64 reference sequences; microbatch 1; accumulation 4.\\[1.5pt]
Execution & Two eight-GPU nodes; separate draft-training and frozen-target-serving processes share devices.\\[1.5pt]
Schedule & Two epochs / 34,432 updates; 17,216 updates per epoch.
\end{tabularx}
\end{recipecard}
The 27B recipe above is separate from the matched offline comparison and training-throughput measurements.

\Needspace{5\baselineskip}
\subsection{Training-Throughput Measurement}\label{app:timing}\label{app:throughput-protocol}
We measure training throughput and execution profiles separately (Section~\ref{sec:training-efficiency}). Here, \emph{native configurations} refer to the method-specific supervision, sampling, backend, and execution settings used in the measurements.

The Qwen3.8-27B / DFlash V2 comparison uses eight high-performance GPUs. Unprofiled timings pool two runs per method over updates 17--32. Supervised tokens count predictions contributing to each method's training objective. The displayed ratios use the underlying measurements rather than ratios of rounded table entries. Offline reference preparation is excluded, and this comparison does not measure time to a matched acceptance threshold.

Figure~\ref{fig:specopd-efficiency}a uses separate profiled windows, updates 17--20, with separate time axes. Grayscale records the fraction of each 50\,ms bin covered by CUDA kernels, collectives, copies, or memset; the reported mean averages over time and GPUs. This measures recorded activity, not SM occupancy, and incomplete capture means white regions do not establish hardware idle time. Blue dashed lines mark update boundaries. Stage shares use aligned host ranges; Draft-OPD's instrumentation does not separately resolve input preparation.

The throughput illustration in Figure~\ref{fig:specopd-efficiency}b uses rectangle width for updates/s, height for supervised tokens/update, and area for supervised tokens/s. The blue rectangle includes the overlaid gray region. These geometric encodings summarize the measured training throughput.

\Needspace{5\baselineskip}
\subsection{Comparison Settings}\label{app:controls}
The compared training pipelines use their respective supervision and execution settings. Training selector support and serving sampling filters are specified separately. We report acceptance and wall-time throughput as separate metrics.

\FloatBarrier

\Needspace{6\baselineskip}
\section{Method Details}\label{app:method-details}
\subsection{Conditioning and Prediction-Row Readout}\label{app:toc-subsection-9}
\paragraph{The Sampled Predecessor Changes the Learning Problem.}
In the running example, a branch starts after \code{a} and samples \code{short}, whereas the reference continues with \code{quiet}. To score the next branch token \code{evening}, the target must read \code{Enjoy a short}, not \code{Enjoy a quiet}. The two histories can induce different target distributions even when the drafter assigns the same probability to the next token.

Divergence from the reference is not necessarily an error: sampling can produce different valid continuations. The first token is compared with the target at the anchor, and later tokens are compared at their sampled predecessor histories. A serving verifier could reject \code{short}, so the latter is a sampled counterfactual history, not necessarily an accepted serving history. Our adapter supervises all retained branch positions; prefix-overlap credit is not a hard rejection mask. Whether supervision beyond a would-be rejection helps requires comparison with a baseline retaining only accepted positions and the first rejection.

\FloatBarrier

\paragraph{Prediction-Row Readout.}
Parallel scoring must preserve the same target queries as scoring each branch separately. We pack the histories into a proposal tree with one shared reference trunk and private branch rows. Branch $b$ can read the trunk only through $a_b$ and then its own predecessors---never future reference tokens or sibling branches. Figure~\ref{fig:tree} specifies this visibility directly as a target attention mask.

Two mappings enforce this history. \emph{Semantic positions} assign token $i$ the rotary position embedding (RoPE) index $a_b+i$, independent of its packed offset. A \emph{prediction-row map} scores the first token from its anchor row and each later token from its predecessor row. Reading a token's own row would leak the answer into its prediction. This is the target scoring mask, not the draft's within-block bidirectional attention.

The target returns only these selected hidden rows; the trainer reconstructs logits with the frozen LM head. This reduces transfer width without restricting teacher vocabulary support. With cached anchor states, scoring $W$ tokens requires only $W-1$ new predecessor rows: the last sampled token has no later prediction to supply.

In Figure~\ref{fig:tree}, the branch at \code{a} samples \code{short}, \code{evening}, and \code{walk}. The anchor row scores \code{short}; the \code{short} row scores \code{evening}; and the \code{evening} row scores \code{walk}. The final sample needs no forward. A plain causal mask over packed offsets would expose future trunk tokens to the branch. Ancestry prevents future-context leakage; predecessor-row readout prevents answer leakage.

\FloatBarrier

\Needspace{5\baselineskip}
\subsection{Acceptance Surrogate and Its Scope}\label{app:toc-subsection-10}
\paragraph{Distribution Overlap and Acceptance.}
Following E2E-TV~\cite{bebop}, let $p_i(\cdot\mid h_i)$ be the target distribution and $q_i(\cdot\mid h_i)$ the proposal distribution at a common history $h_i$. For exact speculative rejection sampling \cite{leviathan,sps2}, a sampled draft token $y_i\sim q_i$ is accepted with probability
\begin{equation}
r_i(y_i,h_i)=\min\left(1,\frac{p_i(y_i\mid h_i)}{q_i(y_i\mid h_i)}\right).
\label{eq:reject}
\end{equation}
Its single-position expectation at a fixed history is
\begin{equation}
\alpha_i(h_i)=\E_{y_i\sim q_i}[r_i]
=\sum_v\min\{q_i(v\mid h_i),p_i(v\mid h_i)\}
=1-\TV(q_i,p_i).
\label{eq:overlap}
\end{equation}
Thus acceptance depends on distribution overlap at a shared history, not reference-token argmax accuracy.

Let $S_i$ be the draft's candidate support. Because $q_i$ is zero outside $S_i$, its total variation distance from the fully normalized target is
\begin{equation}
d_{b,i}=\frac{1}{2}\left[
\sum_{v\in S_i}|q_{b,i}(v)-p_{b,i}(v)|
+1-\sum_{v\in S_i}p_{b,i}(v)\right].
\label{eq:tvd}
\end{equation}
The second term retains teacher mass outside the draft support. Renormalizing $p$ over $S_i$ would remove that mass and change the objective. Distribution comparisons use \texttt{FP32} softmax and reductions, even with \texttt{BF16} hidden states and weights.

For example, let $q=(0.5,0.5,0)$ and $p=(0.4,0.4,0.2)$ over $(X,Y,Z)$. Their TVD is $0.2$ and overlap $0.8$. Renormalizing $p$ over $\{X,Y\}$ instead gives a false perfect match, despite the drafter's inability to propose $Z$.

For $B$ retained branches, define $\alpha_{b,i}=1-d_{b,i}$. Equation~\ref{eq:loss} can equivalently be written as $\mathcal L=N/C$, where $N=\sum_{b=1}^{B}\sum_{j=1}^{W}(1-\prod_{i=1}^{j}\alpha_{b,i})$ and $C=BW$ are the loss numerator and scored-position count, respectively.
The prefix product connects an early mismatch to every later prefix. For one branch,
\begin{equation}
\frac{\partial\mathcal{L}}{\partial d_k}
=\frac{1}{W}\sum_{j=k}^{W}\prod_{\substack{i=1\\i\ne k}}^{j}(1-d_i).
\end{equation}
Products remain differentiable rather than acting as detached weights. Improving an early overlap can protect several downstream terms; this does not imply a universal ordering of gradient magnitudes.

\paragraph{Example: Equal Token-Wise Error, Unequal Prefix Loss.}
Consider $A=(0.9,0.6,0.9)$ with $B=(0.6,0.9,0.9)$. Both have mean TVD $0.2$, but their prefix products are $(0.9,0.54,0.486)$ and $(0.6,0.54,0.486)$, giving losses $0.358$ and $0.458$. The objective distinguishes where the mismatch occurs. At $A$, the TVD derivatives are $(0.7133,0.5700,0.1800)$: the first overlap affects three terms, the last only one.

\paragraph{Scope for Conditional Drafters.}
Equation~\ref{eq:overlap} is exact at a fixed history. Conditional proposals couple later histories to earlier tokens and acceptance events, so
\begin{equation}
\Pr(A_1\cap\cdots\cap A_j)
=\prod_{i=1}^{j}\Pr(A_i\mid A_1,\ldots,A_{i-1}),
\end{equation}
For conditional drafters, the sampled-branch overlap product need not equal this joint probability. We therefore use E2E-TV as an \emph{acceptance-length surrogate}, rather than an unbiased estimator of accepted length. Gradients differentiate overlap at fixed sampled histories and omit the sampling distribution's score-function term. Appendix~\ref{app:counterexample} gives a counterexample; Section~\ref{sec:results} measures actual serving acceptance.

\FloatBarrier

\paragraph{Conditioning Counterexample.}\label{app:counterexample}
Let $q_1=(0.9,0.1)$ and $p_1=(0.1,0.9)$, so $\alpha_1=0.2$. Suppose the second-position overlap is one after the first vocabulary token and zero after the second. Exact joint acceptance is $\sum_y\min(q_1(y),p_1(y))\alpha_2(y)=0.1$, whereas the expected sampled-history product is $\alpha_1\sum_y q_1(y)\alpha_2(y)=0.18$. Fixed-history overlap therefore does not imply exact multi-position survival for conditional proposals. The example limits the surrogate's theoretical interpretation, not the measured acceptance improvement.

\FloatBarrier

\Needspace{6\baselineskip}
\section{Implementation Details}\label{sec:engineering}
Branch independence must survive both model execution and distributed scheduling. We develop a shared-prefix executor that restores private recurrent states, scores indexed branch paths in parallel, and preserves the objective under distributed batching. These mechanisms preserve branch ancestry and objective weighting across execution layouts.

\Needspace{5\baselineskip}
\subsection{Draft Models}\label{app:toc-subsection-11}
For each draft model, we sample branches, obtain frozen-target supervision on the sampled histories, and optimize the E2E-TV surrogate. Sampled token IDs are held fixed during differentiation; the models differ in how they form proposal probabilities.

\paragraph{DFlash.}
The parallel backbone produces block logits, from which we sample proposal tokens after the anchor. The retained student computation graph supplies full-vocabulary probabilities for the E2E-TV loss.

\paragraph{DSpark.}
We reuse parallel block features while sampling tokens sequentially with a Markov bias conditioned on the sampled predecessor. For training, we recompute differentiable proposal probabilities with those same predecessors and compare them with frozen-target distributions along the sampled branch. Our DSpark recipe also trains its confidence head with an auxiliary binary cross-entropy loss targeting detached token-overlap probabilities.

\paragraph{DFlash V2.}
With DFlash V2 at block size eight, we sample $W=7$ branch tokens. The backbone predicts an anchor-plus-mask block in one parallel pass \cite{dflash}. We keep convolution within each anchor block and attention within its configured block/document boundaries. Independent blocks therefore do not mix features.

The selector makes these parallel predictions coherent. At position $i$, it restricts candidates to the unary top-$k$ set $S_i$ and adds a bias conditioned on the sampled predecessor:
\begin{equation}
u_i(v)=\ell_i(v)+\left\langle e_{\rm prev}(y_{i-1})\odot W_h z_i,
e_{\rm next}(v)\right\rangle,\qquad v\in S_i,
\label{eq:selector}
\end{equation}
where $z_i$ is the draft hidden state, $\ell_i$ the unary logit, and $y_0$ the anchor token. Token embeddings $e_{\rm prev}$ and $e_{\rm next}$ interact with projected features $W_hz_i$. The proposal is
\begin{equation}
q_i(v)=\begin{cases}
\displaystyle\frac{\exp(u_i(v)/T_{\rm temp})}{\sum_{w\in S_i}\exp(u_i(w)/T_{\rm temp})},&v\in S_i,\\
0,&v\notin S_i.
\end{cases}
\end{equation}
Here $T_{\rm temp}$ is the proposal temperature. Sampling traverses the candidate sets using the already computed block features. The predecessor changes the selector score; it does not trigger another backbone forward. Proposal and score temperatures are equal and positive, both one in our run. Selector support size 16 is separate from the generation/evaluation filter \code{top\_k=20}.

During backward, top-$k$ candidate identities and sampled token IDs are fixed. Student probabilities remain differentiable, updating both backbone and selector. The target and its distributions are frozen.

\FloatBarrier
\Needspace{5\baselineskip}
\subsection{One Optimization Step}\label{app:toc-subsection-12}
Algorithm~\ref{alg:optimization-step} summarizes one optimizer step. Sampling and loss evaluation use the same draft parameter version; sampled token IDs and candidate identities are held fixed during differentiation.

\begin{algorithm}[!htbp]
\caption{One \method\ optimization step}\label{alg:optimization-step}
\small
\renewcommand{\arraystretch}{1.18}
\newcommand{\algindent}{\hangindent=1.4em\hangafter=1\hspace{0.4em}\textcolor{qwenblue!25}{\vrule width 0.6pt height 1.1ex depth 0.3ex}\hspace{0.8em}}
\begin{tabular}{@{}>{\color{gray}\scriptsize}r@{\hspace{0.8em}}p{0.93\linewidth}@{}}
\rowcolor{qwenblue!4}
\multicolumn{2}{@{\hspace{5pt}}p{0.97\linewidth}@{}}{\textcolor{qwenblue}{Input}\quad Draft $q_\theta$; frozen target $p$; $M$ reference-sequence accumulation slots; branch width $W$.}\\[4pt]
1 & Zero draft gradients.\\
2 & \textbf{for} accumulation slot $m=1,\ldots,M$ \textbf{do}\\
3 & \algindent Extract frozen target features; sample response anchors $\{a_b\}$.\\
4 & \algindent Predict draft blocks and candidate supports with $q_\theta$.\\
5 & \algindent Sample each branch: $y_{b,i}\sim q_\theta(\cdot\mid x_{\leq a_b},y_{b,<i})$.\\
6 & \algindent Pack proposal trees; score prediction predecessors with the frozen target.\\
7 & \algindent Read full-vocabulary $p_{b,i}$ and differentiable $q_{\theta,b,i}$ at the same sampled histories.\\
8 & \algindent Compute overlaps $\alpha_{b,i}=1-\TV(p_{b,i},q_{\theta,b,i})$ and prefix products $\prod_{i=1}^{j}\alpha_{b,i}$.\\
9 & \algindent Form the local loss numerator and position count; backpropagate using Equation~\ref{eq:dp}, scaled by $1/M$.\\
10 & \textbf{end for}\\
11 & Synchronize and clip gradients; update $\theta$ once.\\
12 & Release target-scoring resources.\\[4pt]
\rowcolor{qwenblue!4}
\multicolumn{2}{@{\hspace{5pt}}p{0.97\linewidth}@{}}{\textcolor{qwenblue}{Output}\quad Updated draft parameters $\theta$.}
\end{tabular}
\end{algorithm}

The 27B DFlash V2 recipe uses only the \method\ loss, with unit weight; auxiliary objectives are disabled.

\Needspace{5\baselineskip}
\subsection{Parallel Branch Scoring for Hybrid Recurrent Targets}\label{sec:gdn}
\paragraph{Why a Tree Mask Is Insufficient.}
Hybrid Qwen targets combine attention with GDN layers that compress history into a recurrent state. Masking attention does not prevent a branch from inheriting a future or sibling state. GDN already supports parallel sequence computation \cite{gdn}; our execution problem is to reuse a common prefix while preserving the distinct state of every branch. Independent complete-prefix prefill can batch these queries, but repeats their prefixes.

\paragraph{Restore Once, Then Fork Private Paths.}
The shared-target backend retains trunk attention K/V and \texttt{FP32} GDN checkpoints. With checkpoint interval $C_{\rm GDN}=64$, an anchor after token $a$ restores the state after $\lfloor a/C_{\rm GDN}\rfloor C_{\rm GDN}$ trunk tokens, replays only the residual suffix, then follows its sampled predecessors. Indexed recurrent CTAs load checkpoint tiles into private registers and update them without modifying the shared checkpoints. In Figure~\ref{fig:hybrid-ancestry}, anchors \code{a} and \code{evening} share the checkpoint after \code{Enjoy} but replay different trunk suffixes; neither waits for the other's state update.

\paragraph{Execute Indexed Paths in Parallel.}
We compile branch ancestry into indexed replay paths, checkpoint slots, and convolution windows. Recurrent kernels parallelize over paths, value heads, and value-dimension tiles while retaining ordered updates within each path. The chunk backend gathers replay windows, runs a shared fixed-64 scan from private initial states, and scatters only owned suffix outputs. Each canonical output row has one writer. Thus branches can execute concurrently without duplicating complete prefixes or mixing state; teacher-only kernels require no backward graph. The native-vLLM complete-prefill backend remains an alternative without shared-trunk reuse.

\paragraph{Additional Stateful Components.}
PLE derives each branch's n-gram context from its ancestor token IDs. Its dilated short convolution and the GDN short convolution read the corresponding ancestor windows, including across checkpoint boundaries, while shared history remains read-only. The adapters preserve the native PLE embedding, gating, and residual computations.

\FloatBarrier
\Needspace{5\baselineskip}
\subsection{Deterministic Target Inference}\label{app:deterministic-target}
\paragraph{Stable Supervision across Execution Layouts.}
Small arithmetic differences can change MoE routing or QSA top-$k$ selection and amplify into substantial logit differences. A frozen target can then assign different distributions to the same token history under different execution layouts (Table~\ref{tab:deterministic-target}). This variation can perturb the distillation loss and gradients for computational reasons, making convergence harder to interpret.

\paragraph{Execution Contract and Defaults.}
Building on batch-invariant inference mechanisms in vLLM and SGLang, \method{} enables deterministic target inference by default. For fixed weights, tokens, ancestry, hardware, precision, and backend, we require bitwise-identical full-vocabulary logits across independent full-prefix recomputation (R), cached single or batched branch scoring (C1/CB), and merged-tree execution (T), including fresh full-path repeats. The first draft token uses the anchor's prediction row; subsequent tokens use their predecessors' rows. This option applies to frozen-target scoring; proposal sampling and training semantics remain unchanged.

\paragraph{Consistent Numerical Paths.}
GDN trunk and branch computation share the numerical update scheme. In the chunkwise backend, both invoke the same fixed-64 scan from aligned \texttt{FP32} checkpoints, preserving intermediate casts and true-length masks for partial chunks. With batch invariance enabled, this shared numerical path reduced the 27B logit difference against R to zero. Recurrent execution fixes reduction and fusion boundaries, and GDN gated normalization uses a fixed row layout. QSA fixes score arithmetic, resolves top-$k$ ties by logical group ID, and sorts selected IDs into a canonical attention traversal order. Stable selection thus includes both the selected set and its accumulation order.

\paragraph{Whole-Model Validation.}
The 125B paired comparison uses high-performance GPUs, TP=2/PP=1, \texttt{BF16} weights, \texttt{FP32} states, recurrent/recurrent GDN, indexed QSA with a 2,048-token budget, and eager execution. Two samples with 11 branches provide 41 prediction positions over all 248,320 vocabulary entries. Each path executes three times with fresh trunk computation and caches. Table~\ref{tab:deterministic-target} reports the combined deterministic-mode comparison with fixed backends. Same-path repeats have zero difference in both modes; enabling determinism reduces all cross-layout logit differences to zero.

\begin{table}[!htbp]
\caption{125B target-logit differences across execution layouts. A/B compares candidate A against reference B; KL is the prediction-position mean of $D_{\rm KL}(p_B\Vert p_A)$. The largest listed error occurs at the first draft prediction after the three-token prefix in the first sample.}\label{tab:deterministic-target}
\centering\small
\setlength{\tabcolsep}{7pt}
\renewcommand{\arraystretch}{1.08}
\begin{tabular}{lrrrrrr}\toprule
& \multicolumn{3}{c}{Determinism disabled} & \multicolumn{3}{c}{Determinism enabled}\\
\cmidrule(lr){2-4}\cmidrule(lr){5-7}
Comparison & Max. abs. & Rel. L2 (\%) & Mean KL & Max. abs. & Rel. L2 (\%) & Mean KL\\\midrule
C1/R  & 4.1094 & 14.25 & 0.05840 & 0 & 0 & 0\\
CB/R  & 3.4414 & 14.82 & 0.04958 & 0 & 0 & 0\\
T/R   & 3.7695 & 15.01 & 0.05192 & 0 & 0 & 0\\
CB/C1 & 3.0625 & 11.35 & 0.02743 & 0 & 0 & 0\\
T/CB  & 4.2539 & 14.69 & 0.04382 & 0 & 0 & 0\\\bottomrule
\end{tabular}
\end{table}

An additional QSA pruning test covers 15 branches and 60 prediction positions, with up to 514 visible groups competing for 512 slots. All repeated and cross-path logit differences are zero, including comparisons with native vLLM causal full-prefix execution using the same adapters. These results cover the tested scoring configurations; native cached decoding and CUDA graphs are outside this validation.

\Needspace{5\baselineskip}
\subsection{Objective-Preserving Distributed Normalization}\label{app:toc-subsection-14}
Ranks can have unequal scored-position counts, so averaging their local mean losses changes the objective. For accumulation slot $m$, rank $r$ contributes numerator $N_{r,m}$ and count $C_{r,m}$. We use
\begin{equation}
\mathcal{L}_m=\frac{\sum_r N_{r,m}}{\max(1,\sum_r C_{r,m})},\qquad
\mathcal{L}_{\rm step}=\frac{1}{M}\sum_{m=1}^{M}\mathcal{L}_m.
\label{eq:dp}
\end{equation}
Because distributed gradients are averaged across $R$ ranks, each rank backpropagates $R N_{r,m}/\sum_r C_{r,m}$. Counts stay attached to their original slots: pooling all slots generally differs from averaging their losses.

The implementation reduces counts independently of target completion order. Merged rows retain sample lengths, proposal RNG identities, and prescribed loss order; balancing preserves logical-rank anchor draws. Empty proposal sets contribute finite zeros. Scheduling therefore cannot silently reweight the data.

\paragraph{Example: Why Rank-Local Means Are Insufficient.}
For $(N_0,C_0)=(0.1,1)$ and $(N_1,C_1)=(2.7,9)$, the local-loss average is $0.2$, but the global-count loss is $2.8/10=0.28$. Two-rank gradient averaging needs local quantities $2N_r/10$, or $0.02$ and $0.54$. Their average recovers $0.28$, regardless of which sample finishes first.

\FloatBarrier

\Needspace{6\baselineskip}
\section{Cost and Complexity Analysis}\label{sec:motivation}\label{app:work}
We distinguish the dependency depth of online collection, target-scoring work, and amortized update cost. These analytical quantities explain possible savings; they are not measured speedups.

\Needspace{5\baselineskip}
\subsection{Collection and Amortized Update Cost}\label{sec:cost}
\paragraph{Online Trajectory Collection.}
An online trajectory of length $T_{\rm gen}$ requires approximately $R_{\rm rollout}=T_{\rm gen}/\bar\tau$ dependent draft--verify rounds, where $\bar\tau$ is mean tokens advanced per verification, including the bonus token. With mean round time $\bar c_{\rm round}$, a non-overlapped single-trajectory cost is
\begin{equation}
C_{\rm traj}\approx C_{\rm prefill}+\frac{T_{\rm gen}}{\bar\tau}\bar c_{\rm round}.
\label{eq:rollout-cost}
\end{equation}
Here $C_{\rm prefill}$ prepares the initial context, while round time covers drafting, verification, and prefix advancement. With $T_{\rm gen}=2{,}048$ and $\bar\tau=4$, collection has about 512 dependent rounds. \method{} replaces this collection stage with branch sampling at reference anchors; trunk extraction, scoring, and learning remain.

\paragraph{Shared-Prefix Work.}
Let $L$ be trunk length, $K$ the branch count, and $\ell_b\leq L$ the prefix length at anchor $b$. A width-$W$ branch reads its first prediction from the anchor, requiring $P=W-1$ new predecessor rows. Compared with an \emph{uncached complete-prefix scorer}, shared execution avoids repeated prefixes after the common trunk pass:
\begin{equation}
N_{\rm independent}=L+\sum_{b=1}^{K}\ell_b+KP,
\qquad N_{\rm shared}=L+KP.
\label{eq:row-work}
\end{equation}
At $L=2{,}048$, $W=7$, and $K=512$ uniformly placed anchors, expected projection/MLP rows fall from 529,664 to 5,120 ($103.45\times$). Cached baselines can already reuse some prefix work. Attention and recurrence retain additional prefix interactions, quantified below and in Figure~\ref{fig:efficiency-model}a.

\paragraph{End-to-End Training Cost.}
Let $C_{\rm rollout}$ be elapsed time for an online collection batch reused over $H$ updates, and $C_{\rm reference}$ the reference-corpus construction time amortized over $U$ updates. Let $t_{\rm learn,rollout}$ and $t_{\rm learn,\text{\method}}$ be average update times excluding these collection costs. For non-overlapped collection and learning, the workflow speedup is
\begin{equation}
S_{\rm workflow}=\frac{C_{\rm rollout}/H+t_{\rm learn,rollout}}
{C_{\rm reference}/U+t_{\rm learn,\text{\method}}}.
\label{eq:workflow-cost}
\end{equation}
The model accounts for reference preparation, collection-batch concurrency, and the distinct learning times. Removing online collection changes the first term; shared-prefix execution reduces work in the second. For overlapped pipelines, elapsed time follows the joint critical path. Appendix~\ref{app:timing} gives our measurement procedure.

\FloatBarrier
\Needspace{5\baselineskip}
\subsection{Attention and Recurrent Work}\label{app:toc-subsection-16}
Let $A(n)=n(n+1)/2$ count causal query--key pairs. With the notation of Equation~\ref{eq:row-work}, we count attention and GDN work separately from projection rows.

Branch queries must still attend to their anchor prefixes:
\begin{align}
A_{\rm independent}&=A(L)+\sum_b A(\ell_b+P),\\
A_{\rm shared}&=A(L)+\sum_b\left[P\ell_b+A(P)\right].
\label{eq:attention-work}
\end{align}
Shared execution removes $\sum_b A(\ell_b)$ repeated prefix pairs, but retains branch--prefix interactions. Kernel overhead can reduce the benefit; the flash-attention lowering still gathers ancestor KV into temporary buffers.

For a hybrid model's Gated DeltaNet (GDN) layer with checkpoints every $C_{\rm GDN}$ trunk tokens, branch $b$ restores the state after $C_{\rm GDN}\lfloor\ell_b/C_{\rm GDN}\rfloor$ tokens and replays $r_b=\ell_b\bmod C_{\rm GDN}$ trunk updates. Its logical recurrence count is
\begin{equation}
N_{\rm GDN}=L+KP+\sum_b r_b\ \leq\ L+KP+K(C_{\rm GDN}-1).
\label{eq:gdn-work}
\end{equation}
Replay uses stored projected features rather than full-model forwards. The count excludes checkpoint storage, fixed-tile execution, and state-copy overhead.

\FloatBarrier
\Needspace{7\baselineskip}
\subsection{Conditional Wall-Time Sensitivity}\label{app:toc-subsection-17}
\begin{figure}[!htbp]
\centering
\includegraphics[width=\textwidth]{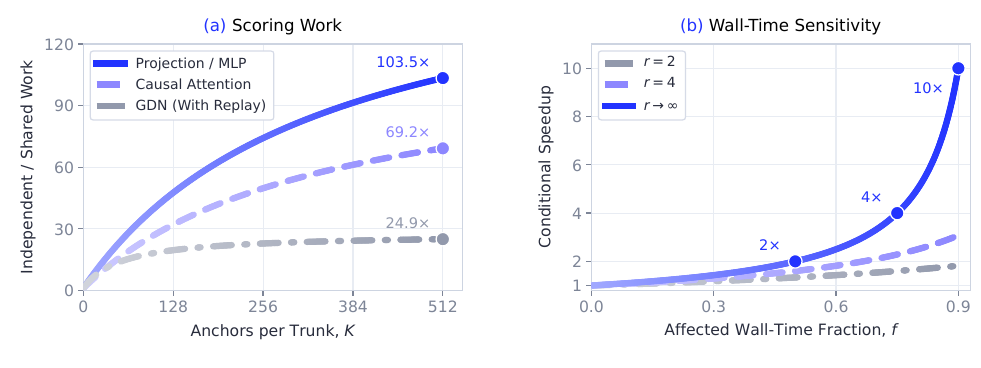}
\caption{Analytical work and conditional wall-time gains. (a) Uncached independent versus shared-prefix scoring with $L=2{,}048$, $W=7$ ($P=6$), $C_{\rm GDN}=64$, and uniformly placed anchors. GDN counts include suffix replay; attention counts causal pairs. (b) Equation~\ref{eq:amdahl} for component acceleration $r$ and no added overhead ($\delta=0$).}
\label{fig:efficiency-model}
\end{figure}

If a fraction $f$ of baseline wall time can be accelerated by factor $r$, while the remaining work is unchanged, the conditional end-to-end gain is
\begin{equation}
S=\frac{1}{(1-f)+f/r+\delta},
\label{eq:amdahl}
\end{equation}
where $\delta$ is added overhead normalized to baseline time. Even removing the affected component entirely ($r\rightarrow\infty$, $\delta=0$) gives only $2\times$ at $f=0.5$ or $4\times$ at $f=0.75$. Actual time savings depend on the measured cost breakdown. Figure~\ref{fig:efficiency-model}b illustrates this dependence with zero added overhead.

\FloatBarrier
\section{Detailed Related Work}
\label{app:detailed-related-work}
\label{app:detailed-speculative-decoding}

\paragraph{Parallel Drafting for SD.}
Speculative decoding verifies inexpensive proposals in parallel~\cite{leviathan,sps2}. Auxiliary heads, target features, multi-token prediction, and self-speculation improve proposal generation~\cite{medusa,eagle,eagle3,mtp,deepseekv3,zhang2024draft}. SWIFT adapts layer skipping online without training~\cite{swift}; KNN-SSD retrieves input-dependent layer sets~\cite{knnssd}. Candidate trees support joint verification~\cite{specinfer,eagle2}, while TALON and ECHO allocate candidates using confidence~\cite{liu2026talon,hu2026echo}. D-cut jointly prunes draft depth across batched requests using confidence and verification cost~\cite{liu2026dcut}. Retrieval supplies candidates from datastores, cached transitions, or target logits~\cite{rest,tokenrecycling,liu2025logitspec}; GRAFT combines retrieval and pruning~\cite{shen2026graft}. Other systems overlap drafting and verification or use rollback-aware branches~\cite{pearl,specbranch,double}. Parallel and semi-autoregressive architectures reduce serial proposal cost~\cite{falcon,pard,peagle,dflare,domino}, including block diffusion in DFlash and lightweight sequential modeling over a parallel backbone in DSpark~\cite{dflash,dspark}. These inference mechanisms can accompany drafter learning when proposal distributions and verification rules are preserved.

\paragraph{On-Policy Draft Training and Acceptance Objectives.}
Fixed reference responses differ from drafter-generated histories. Generalized on-policy distillation supervises student-induced contexts~\cite{gkd}; DistillSpec studies draft-generated data and decoding-aware divergences~\cite{distillspec}; Draft-OPD uses target-assisted rollouts and error-state replay~\cite{draftopd}. EAGLE-3 also accounts for generated draft inputs during training~\cite{eagle3}. Because acceptance stops at the first rejection, tokenwise accuracy alone does not determine progress. BeBop's E2E-TV couples distribution overlap across prefixes~\cite{bebop}, while Expected Decoding Rounds models progress across verification rounds~\cite{edr}. \method{} supplies fresh branch samples and history-consistent target distributions to E2E-TV through parallel collection and scoring, complementing efforts to reduce training memory~\cite{loram} and communication~\cite{edgc}. Its supervision is on-policy within each branch; reference prefixes remain distinct from unrestricted rollout contexts. Collection policy, scoring, and objective choice are separate experimental factors.

\paragraph{Shared Computation across Histories.}
SpecInfer and Medusa jointly score candidate paths through tree attention~\cite{specinfer,medusa}. Shared ancestors permit reuse, but divergent tokens require separate histories. KVShot studies a complementary representation issue: target KV preserves context that hidden states may discard, while introducing query-estimation and gradient-sparsity challenges~\cite{liu2026kvshot}. History preservation also depends on the target architecture: GQA stores token KV~\cite{gqa}, QSA selects blocks through compressed indexer keys~\cite{qsa}, and GDN maintains recurrent state~\cite{gdn}. Branch-parallel supervision therefore combines tree masks, branch-consistent indexer groups, and indexed state restoration and replay. Training additionally requires numerically consistent supervision at every sampled position, beyond identifying an accepted inference prefix. At the model level, HMI shares hierarchical components across tenants~\cite{hmi}, while FloE reduces expert-loading costs through compression and sparse prediction~\cite{floe}; these mechanisms address model memory and movement rather than branch-history consistency.

\paragraph{Speculative Decoding for RL Rollouts.}
RL post-training couples repeated trajectory generation with an evolving policy. SPEC-RL reuses earlier trajectory segments~\cite{liu2025specrl}; SRT refreshes per-prompt continuation trees~\cite{chang2026srt}. Integrated systems support synchronous and asynchronous speculative rollouts~\cite{iso2026rlspec}, and BeBop combines MTP with rejection sampling~\cite{bebop}. TailSieve isolates long-tail requests and allocates route-specialized replicas using MTP or DFlash~\cite{xu2026tailsieve}. Branch-parallel supervision could help refresh learned drafters as the policy changes; this would require version-consistent target supervision and update costs amortized by rollout savings.

\paragraph{Multimodal Speculative Decoding.}
Visual inputs introduce distinct encoding, prefilling, and decoding bottlenecks~\cite{vlmefficiency}. SpecVLM prunes draft-side video tokens while retaining full-target verification~\cite{ji2025specvlm}; ParallelVLM combines visual-alignment-aware pruning with parallel draft--verify execution~\cite{kong2026parallelvlm}. LVSpec instead relaxes verification away from visual anchors, changing the criterion rather than preserving exact target-distribution equivalence~\cite{ji2026lvspec}. Complementary methods reduce conditioning costs through token selection in OmniDelta~\cite{huang2026omnidelta} or head-specific cache compression for multimodal LMs in HybridKV and AR video diffusion in Forcing-KV~\cite{hybridkv,ji2026forcingkv}. Shared visual prefixes could support branch-parallel drafter training, but this extension remains unevaluated. It would require consistent visual features, positions, and cache semantics across branches; exactness relative to an uncompressed target would also require preserving its verification distribution.

\clearpage
\includepdf[pages=1,pagecommand={\thispagestyle{empty}},fitpaper=true]{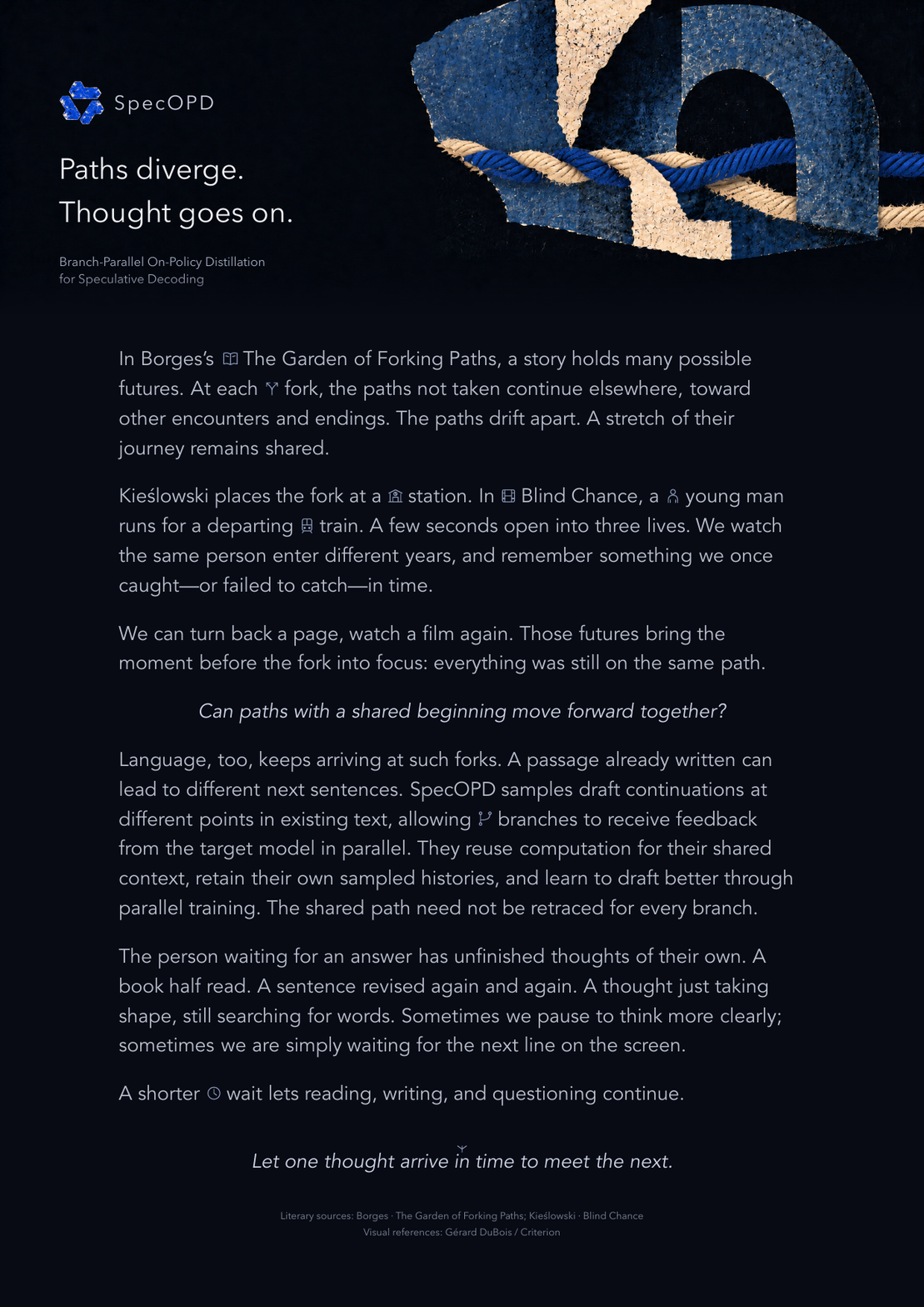}
\end{document}